\documentclass[times,twocolumn,final,authoryear]{elsarticle}

\usepackage{framed,multirow}
\usepackage{subcaption}
\usepackage{amssymb}
\usepackage{latexsym}
\usepackage{amsmath,amsfonts,bm}
\usepackage{multirow}
\usepackage{makecell}

\usepackage{mdwlist} 
\usepackage{enumitem}

\usepackage{url}
\usepackage[usenames,dvipsnames]{xcolor}
\definecolor{newcolor}{rgb}{.8,.349,.1}

\usepackage{glossaries}
\makeglossaries

\newacronym{PSI}{PSI}{Photonic Systems Integration}
\newacronym{ACFR}{ACFR}{Australian Centre for Field Robotics}
\newacronym{CRIS}{CRIS}{Centre for Robotics and Intelligent Systems}
\newacronym{ACRA}{ACRA}{the Australasian Conference on Robotics and Automation}
\newacronym{ACRV}{ACRV}{Australian Centre for Robotic Vision}
\newacronym{USyd}{USyd}{the University of Sydney}
\newacronym{UQ}{UQ}{the University of Queensland}
\newacronym{QUT}{QUT}{the Queensland University of Technology}
\newacronym{UCSD}{UCSD}{the University of California, San Diego}
\newacronym{ANU}{ANU}{Australia National University}
\newacronym{IMOS}{IMOS}{the Integrated Marine Observation System}
\newacronym{URI}{URI}{the University of Rhode Island}
\newacronym{WHOI}{WHOI}{Woods Hole Oceanographic Institution}

\newacronym{NSF}{NSF}{National Science Foundation}
\newacronym{LIEF}{LIEF}{Linkage Infrastructure, Equipment and Facilities}

\newacronym{ICCP}{ICCP}{the International Conference on Computational Photography}
\newacronym{CVPR}{CVPR}{Computer Vision and Pattern Recognition}
\newacronym{TIP}{TIP}{Transactions on Image Processing}
\newacronym{TSP}{TSP}{Transactions on Signal Processing}
\newacronym{JFR}{JFR}{the Journal of Field Robotics}
\newacronym{ISCAS}{ISCAS}{International Symposium on Circuits and Systems}
\newacronym{TOG}{TOG}{Transactions on Graphics}
\newacronym{ICRA}{ICRA}{International Conference on Robotics and Automation}
\newacronym{IROS}{IROS}{Intelligent Robots and Systems}
\newacronym{RA-L}{RA-L}{Robotics and Automation Letters}

\newacronym{AUV}{AUV}{autonomous underwater vehicle}
\newacronym{UAV}{UAV}{unmanned aerial vehicle}
\newacronym{USV}{USV}{unmanned surface vehicle}
\newacronym{UGV}{UGV}{unmanned ground vehicle}
\newacronym{GPS}{GPS}{global positioning system}
\newacronym{SLAM}{SLAM}{simultaneous localisation and mapping}
\newacronym{SfM}{SfM}{structure from motion}
\newacronym{AR}{AR}{augmented reality}
\newacronym{VR}{VR}{virtual reality}
\newacronym{MR}{MR}{mixed reality}
\newacronym{CNN}{CNN}{convolutional neural network}
\newacronym{GAN}{GAN}{generative adversarial network}
\newacronym{DNN}{DNN}{deep neural network}
\newacronym{IMU}{IMU}{inertial measurement unit}
\newacronym{TOF}{TOF}{time of flight}
\newacronym{GSD}{GSD}{ground spatial distance}

\newacronym{MDSP}{MDSP}{multi-dimensional signal processing}
\newacronym{ROS}{ROS}{region of support}
\newacronym{DOF}{DOF}{degree-of-freedom}
\newacronym{RMS}{RMS}{root mean square}
\newacronym{RMSE}{RMSE}{root mean squared error}
\newacronym{SNR}{SNR}{signal-to-noise ratio}
\newacronym{CNR}{CNR}{contrast-to-noise ratio}
\newacronym{PCA}{PCA}{principal component analysis}
\newacronym{MSE}{MSE}{mean squared error}
\newacronym{MAE}{MAE}{mean absolute error}
\newacronym{IoU}{IoU}{intersection over union}
\newacronym{mIoU}{mIoU}{mean intersection over union}

\newacronym{FIR}{FIR}{finite impulse response}
\newacronym{IIR}{IIR}{infinite impulse response}
\newacronym{DFT}{DFT}{discrete Fourier transform}
\newacronym{FFT}{FFT}{fast Fourier transform}
\newacronym{PSNR}{PSNR}{peak signal-to-noise ratio}
\newacronym{SSIM}{SSIM}{structural similarity index}
\newacronym{MS-SSIM}{MS-SSIM}{multi-scale structural similarity index}
\newacronym{FPGA}{FPGA}{field programmable gate array}
\newacronym{GPU}{GPU}{graphics processing unit}
\newacronym{ASIC}{ASIC}{application-specific integrated circuit}
\newacronym{BW}{BW}{bandwidth}

\newacronym{PSF}{PSF}{point spread function}
\newacronym{SPAD}{SPAD}{single-photon avalanche diode}
\newacronym{FOV}{FOV}{field of view}
\newacronym{BRDF}{BRDF}{bidirectional reflectance distribution function}
\newacronym{FWHM}{FWHM}{full width at half maximum}
\newacronym{LF}{LF}{light field}
\newacronym{2pp}{2pp}{two-plane parameterization}
\newacronym{MLA}{MLA}{microlens array}

\newacronym{RANSAC}{RANSAC}{random sampling and consensus}
\newacronym{DoG}{DoG}{difference of Gaussian}
\newacronym{SIFT}{SIFT}{scale invariant feature transform}

\newacronym{NIR}{NIR}{near-infrared}  
\newacronym{HOG}{HOG}{hisogram of oriented gradient}
\newacronym{SVM}{SVM}{support vector machine}
\newacronym{BoW}{BoW}{bag of words}
\usepackage{verbatim} 

\usepackage[bookmarks=true,hidelinks]{hyperref}

\journal{Computer Vision and Image Understanding}

\begin{document}

\thispagestyle{empty}

\clearpage

\ifpreprint
  \setcounter{page}{1}
\else
  \setcounter{page}{1}
\fi

\begin{frontmatter}

\title{Semantically Accurate Super-Resolution Generative Adversarial Networks}

\author[1]{Tristan {Frizza}} 
\author[2]{Donald G. {Dansereau}\texorpdfstring{\corref{cor1}}{}}
\cortext[cor1]{Corresponding author: Rose St Bldg J04, University of Sydney, Australia}
\ead{donald.dansereau@sydney.edu.au}
\author[1]{Nagita Mehr {Seresht}}  
\author[1]{Michael {Bewley}}

\address[1]{Nearmap, Level 4 Tower One, International Towers, Barangaroo NSW 2000}
\address[2]{Australian Centre for Field Robotics, University of Sydney, NSW 2006, Australia}

\begin{abstract}
This work addresses the problems of semantic segmentation and image super-resolution by jointly considering the performance of both in training a Generative Adversarial Network (GAN). We propose a novel architecture and domain-specific feature loss, allowing super-resolution to operate as a pre-processing step to increase the performance of downstream computer vision tasks, specifically semantic segmentation. We demonstrate this approach using Nearmap's aerial imagery dataset which covers hundreds of urban areas at 5-7~cm per pixel resolution. We show the proposed approach improves perceived image quality as well as quantitative segmentation accuracy across all prediction classes, yielding an average accuracy improvement of 11.8\% and 108\% at 4$\times$ and 32$\times$ super-resolution, compared with state-of-the art single-network methods. This work demonstrates that jointly considering image-based and task-specific losses can improve the performance of both, and advances the state-of-the-art in semantic-aware super-resolution of aerial imagery. 
\end{abstract}

\begin{keyword}
super-resolution \sep semantic segmentation \sep generative adversarial networks \sep multi-modal learning
\end{keyword}

\end{frontmatter}



\section{Introduction}

Image super-resolution is a well studied but challenging task in which low-resolution imagery is to be upsampled to a high-fidelity, higher resolution image. While traditional methods draw on decades of image and signal-processing knowledge \citep{farsiu2004advances}, these fail to exploit the rich \emph{prior} knowledge we have of what realistic scenes look like. Such knowledge is difficult or impossible to capture by traditional means, but is accessible via learning-based approaches \citep{anwar2020deep}, allowing these to achieve much greater levels of fidelity and more aggressive upsampling rates. 

Most recently, \gls{GAN}-based super-resolution architectures \citep{goodfellow2014generative} have demonstrated state-of-art performance by pairing a generator and adversarial discriminator \citep{wang2020ultra}. In this scheme, the discriminator learns to distinguish synthetic and captured imagery, while the generator competitively learns to fool it with more realistic generated content. 

However, in the context of applications in which semantic meaning is as important as visual realism, \glspl{GAN} risk filling in details that are not semantically faithful. This is of utmost importance in a remote sensing context where we wish to improve resolution without compromising semantic accuracy.

\begin{figure}
\captionsetup[subfloat]{justification=centering}
\centering
\includegraphics[width=0.32\linewidth]{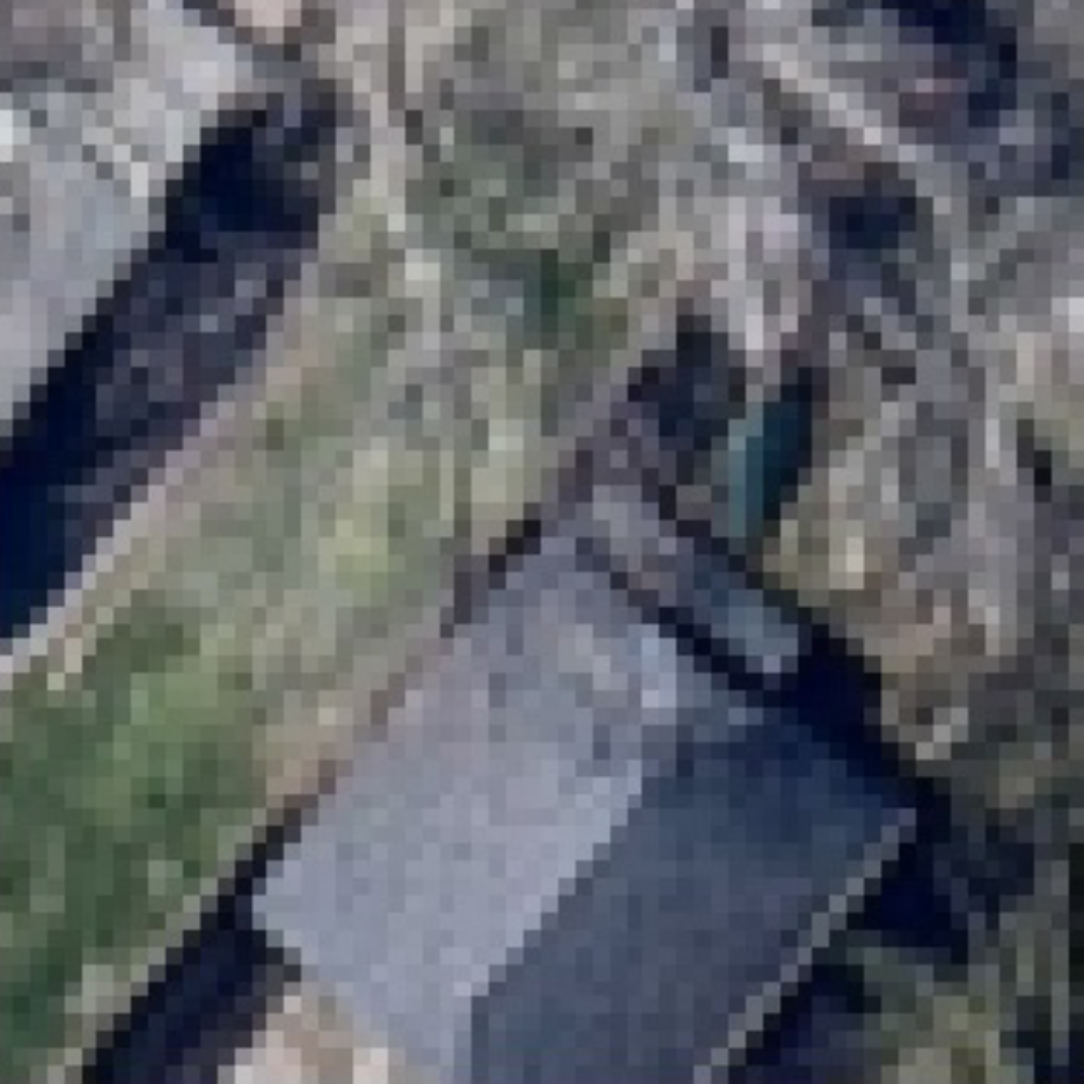}%
\hspace*{.1cm}
\includegraphics[width=0.32\linewidth]{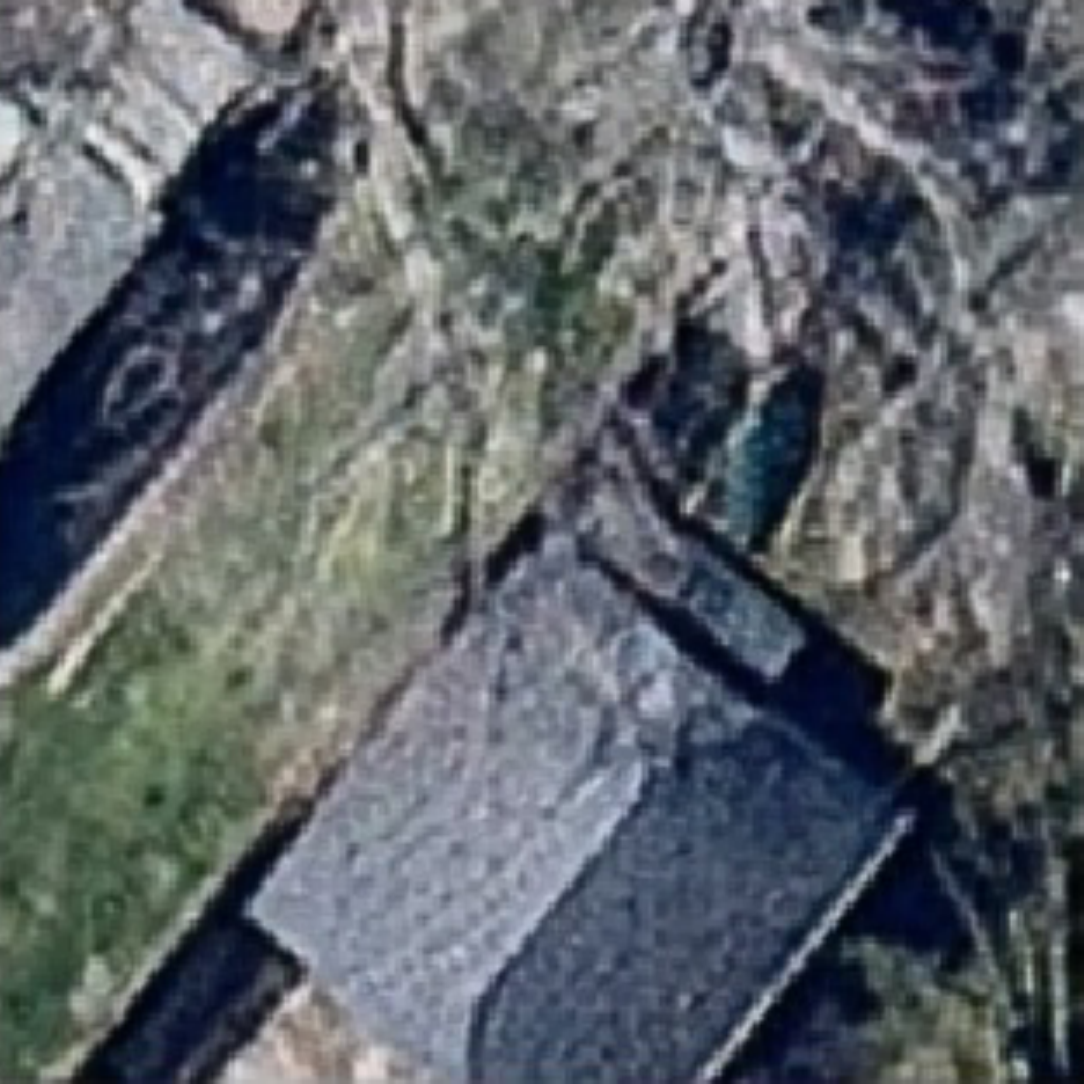}%
\hspace*{.1cm}
\includegraphics[width=0.32\linewidth]{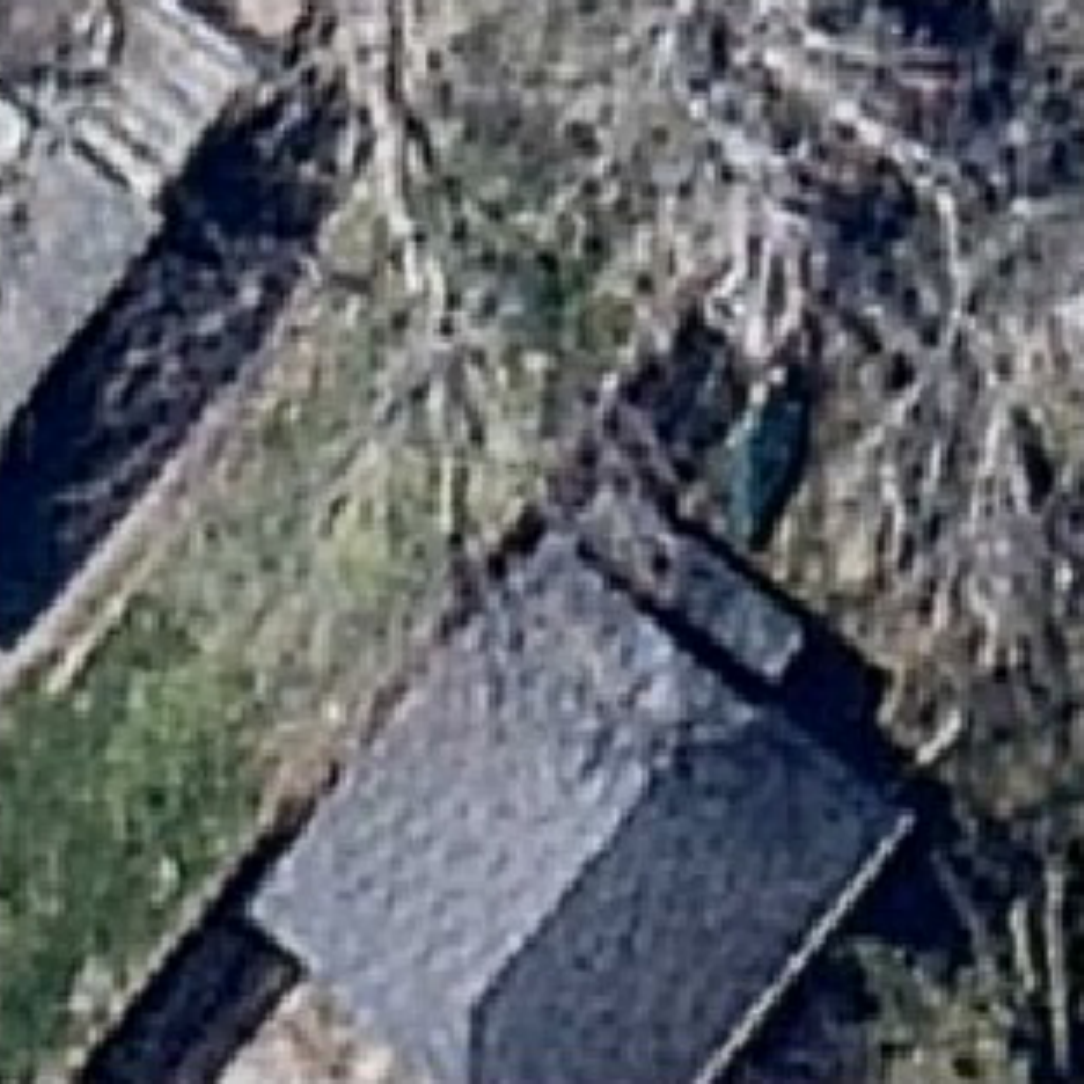}%
\vspace*{-0.2cm}

\centering
\subfloat[][Input]{\includegraphics[width=0.32\linewidth]{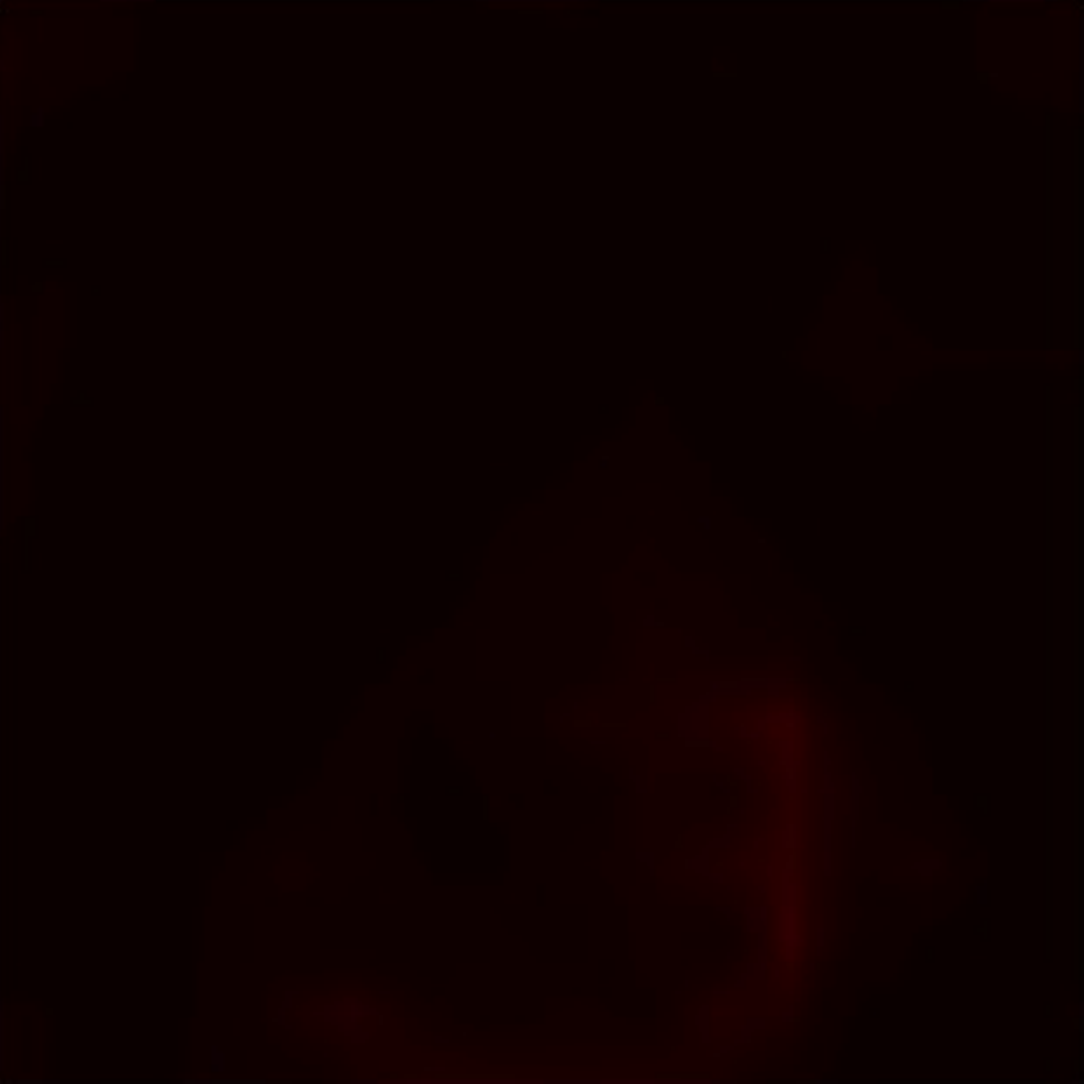} \label{fig:opening_lr}}
\hspace*{.025cm}
\subfloat[][Super-resolved]{\includegraphics[width=0.32\linewidth]{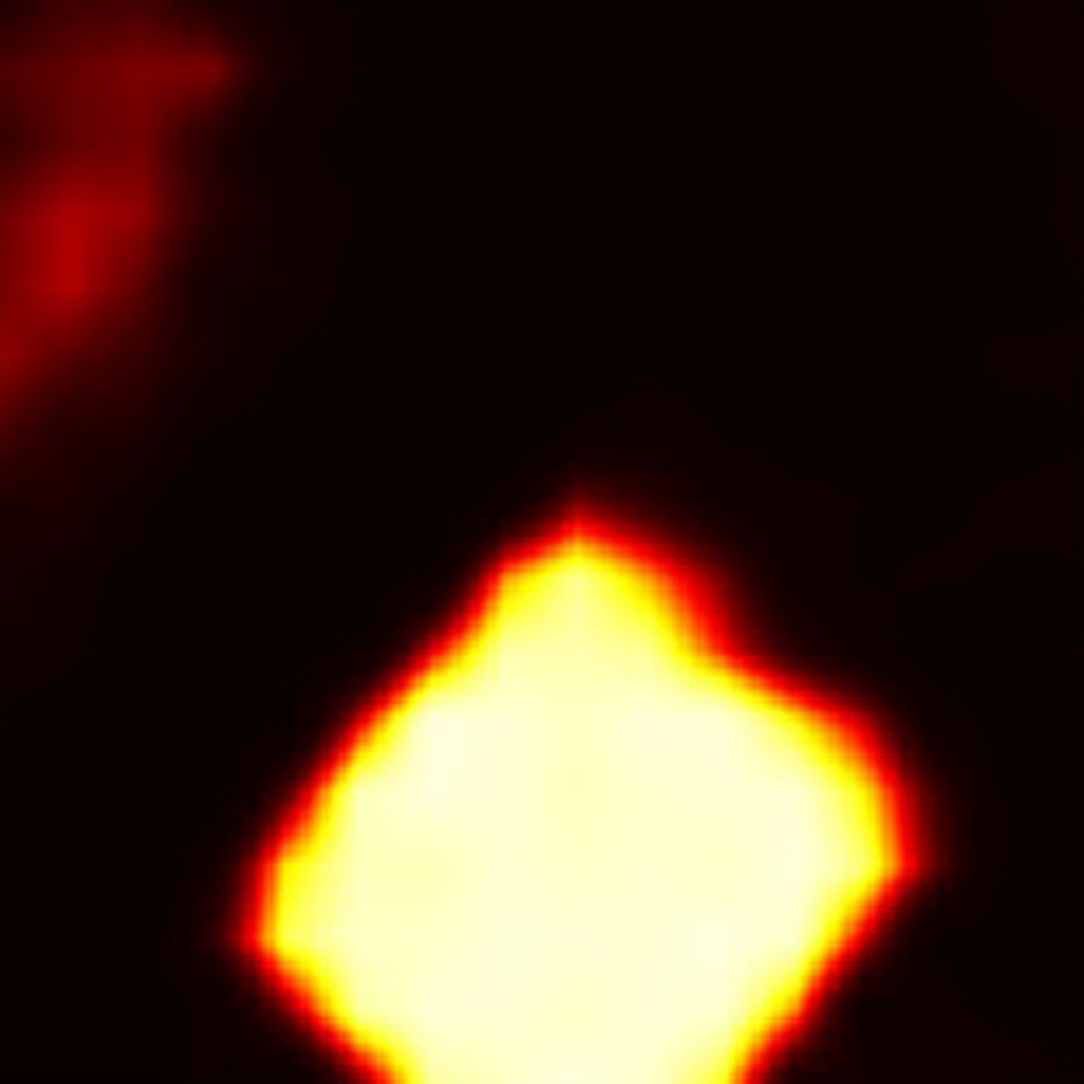} \label{fig:opening_sr}} 
\hspace*{.025cm}
\subfloat[][Target]{\includegraphics[width=0.32\linewidth]{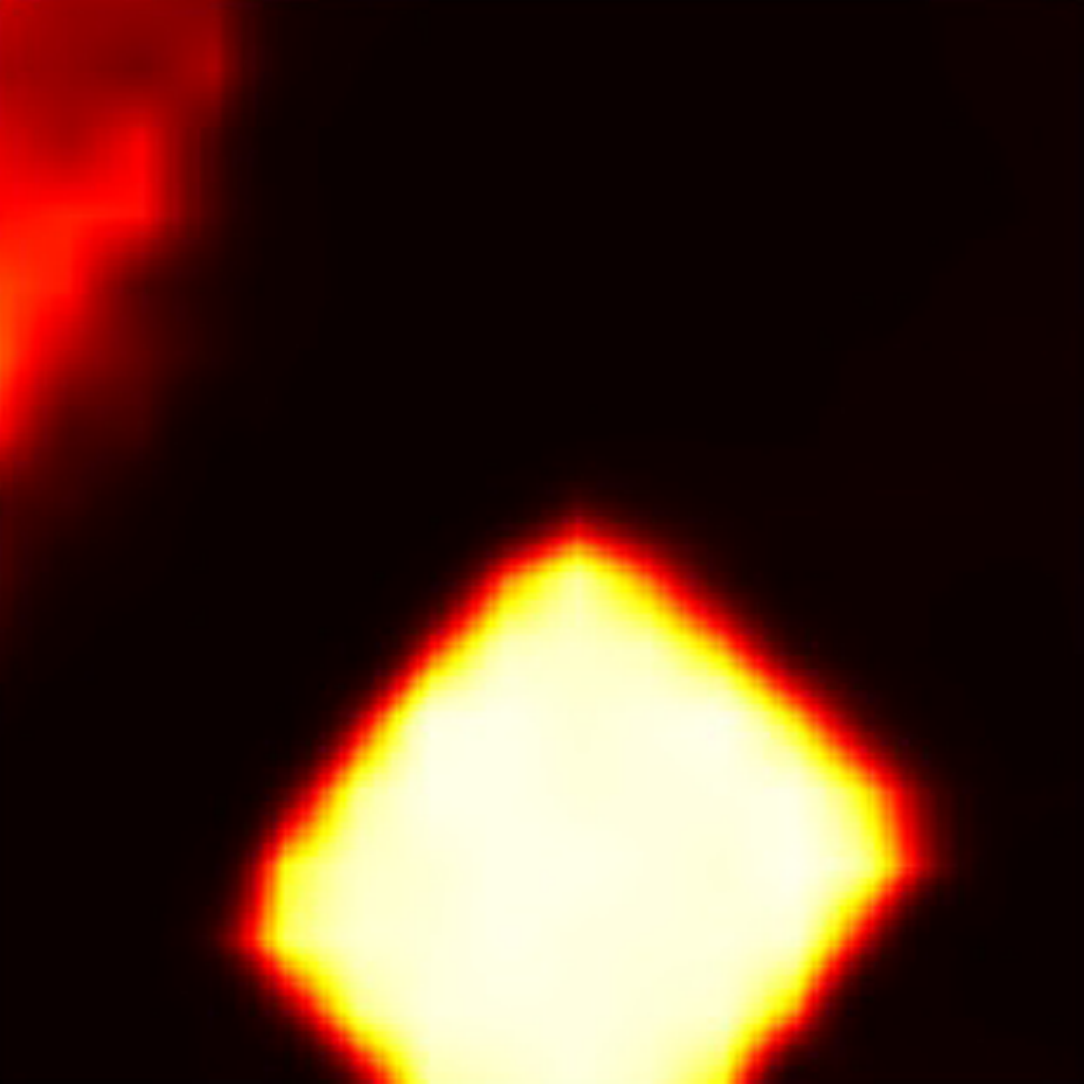} \label{fig:opening_hr}} 
\caption{(a) Using conventional segmentation approaches, low-resolution aerial imagery (top) yields poor semantic segmentation predictions (bottom). In this work we present a \gls{GAN}-based super-resolution strategy that improves both image resolution and semantic segmentation accuracy, yielding results (b) comparable to imagery captured at 4 times the resolution (c).}
\label{fig:teaser}
\end{figure}

In this work we propose to jointly consider visual and semantic fidelity in training a super-resolution \gls{GAN}. We show that considering both problems at once improves performance at both, as seen in Fig.~\ref{fig:teaser}. The contributions of this work are:

\begin{itemize}[noitemsep,topsep=0pt]
\item We present a \gls{GAN}-based architecture for super-resolving aerial imagery by leveraging semantic label information,
\item We introduce a domain-specific feature loss that jointly optimises detail-oriented perceptual realism as well as semantic accuracy of aerial image features, and
\item We demonstrate state-of-the-art super-resolution and improved semantic segmentation of aerial imagery used in an industrial remote sensing application.
\end{itemize}

This work establishes that a form of multi-modal learning, in which semantics and visual fidelity are jointly addressed, yields superior results at both tasks. We show results for super-resolution in a specific application domain, and anticipate generalisation to both different domains and to applications beyond super-resolution.

The remainder of this paper is organised as follows: in Sec.~\ref{sec_RelateWork} we review related work and indicate how our relates to what has come before. In Sec.~\Ref{sec_Method} we outline our approach, a novel \gls{GAN}-based architecture and loss functions that jointly addresses semantic and visual fidelity in super-resolving aerial imagery. Sec.~\ref{sec_Results} evaluates the method with quantitative and qualitative comparisons to competing methods. Finally, Sec.~\ref{sec_Conclusions} summarises the work and indicates directions for future work.

\section{Related Work}
\label{sec_RelateWork}

This work is concerned with single-image super-resolution. Traditional approaches range in complexity from interpolation \citep{Keys1981} to statistical regression with natural image priors \citep{Kim2010} to sparse coding methods \citep{J.2010ImageRepresentation}. 

Recently, learning-based methods have demonstrated vastly improved super-resolution performance achieved \citep{anwar2020deep, saharia2021image}. Early applications of neural networks such as SRCNN \citep{C.2016ImageNetworks} made use of \glspl{CNN} to learn non-linear mappings from low to high-resolution image spaces. Later improvements such as VDSR \citep{Kim2016AccurateNetworks} saw improved success by increasing network depth. The more recent \gls{CNN} architectures for super-resolution have shown significant advances by using more sophisticated upsampling strategies, e.g.~Laplacian Pyramids \citep{Lai2017FastNetworks}, as well as more efficient layer organisation, e.g.~residual \citep{Ledig2017Photo-realisticNetwork} and dense \citep{ZhangResidualSuper-Resolution} network connections.

Innovations in the choice of loss function for image learning tasks have also yielded significant improvements in the fidelity of super-resolution algorithms. Most notable has been the use of so-called perceptual or feature losses \citep{Johnson2016PerceptualSuper-Resolution} that calculate an image loss with respect to the intermediate activations of an auxiliary VGG classification network, rather than a typical \gls{MSE} loss on the predicted pixels. This technique yields improved reconstruction of the semantic features of an image, without tending toward blurry, pixel-averaged representations typical of an \gls{MSE}-based loss.

Most recently, \glspl{GAN} have proven extremely powerful for modelling high-dimensional image data \citep{goodfellow2014generative, Karras2018ANetworks, wang2020ultra}. These pit a generative model against an adversarial discriminator, jointly optimising such that both generation and discrimination improve through training.  Their formulation as generative models suits the task of super-resolution well as additional detail must be injected into a low-information image. The main strength of \glspl{GAN} is in their use of a learned loss function, parameterised by a separate loss network which is trained adversarially against the original generator. Models such as SRGAN \citep{Ledig2017Photo-realisticNetwork} and ESRGAN \citep{Wang2019ESRGAN:Networks} have been shown to outperform conventional \glspl{CNN} in terms of fine texture detail and perceived perceptual similarity.

\subsection{Super-Resolution in Remote Sensing}

Super-resolution has been an increasingly important problem in remote sensing with important applications in environmental monitoring and global change analysis. Accurate identification, localisation and measurement of objects on the ground is a long-running challenge in remote sensing systems. Satellite imaging, while able to scan large areas of the earth's surface, suffers from low resolution due to imaging from large distances, as well as significant atmospheric distortion. In this work we focus more on high-fidelity overhead aerial imagery that is taken from light aircraft. Rather than spatial resolutions of 30~m typical of LandSat satellite imagery, aerial imaging can reach resolutions in the 5~cm range.

Traditional interpolation, Fourier-domain and Bayesian approaches to super-resolution have found use on aerial imagery \citep{Yang2015RemoteApproaches}. Among more recent works adopting machine-learning approaches, we highlight the work by \cite{Bosch2017Super-ResolutionLearning} that achieves impressive results on both satellite and aerial imagery, using DenseNets \citep{Huang2017DenselyNetworks} and adversarial learning. However, as the authors note, use of an ImageNet pre-trained network for computing feature matching loss exhibits little correlation between features found in natural and overhead images. Furthermore, the authors acknowledge that their model hallucinates unrealistic detail when approaching higher (8x) upsampling factors, alluding to the fact that this could be improved by including coarse semantic results via a conditional \gls{GAN} framework. 

In this work we address these shortcomings of prior work, using a custom-trained aerial image segmentation model to compute feature loss, imbuing the model with semantically relevant feature comparisons, as well as employing a conditional discriminator that ensures the generated images maintain semantic relevance to the original images, without excessive hallucination. With these enhancements we are able to demonstrate meaningful enhancement of resolution at greater upslamping rates, up to 32x.

\subsection{Semantic Segmentation in Remote Sensing}

Semantic image segmentation involves the dense prediction of image region masks and corresponding classes for each pixel in an image \citep{he2017mask}. This has seen widespread adoption in autonomous driving and is gaining attention in remote sensing for localising and labelling important ground-level features like vehicles, buildings and other landmarks. Popular models such as FCN \citep{LongFullySegmentation}, U-Net \citep{Ronneberger2015U-Net:Segmentation} and DeepLab \citep{deeplabv3plus2018} have consistently demonstrated the effectiveness of deep neural networks for this task.

Several recent works have demonstrated the benefit of applying super-resolution to improve object classification in aerial images. \cite{Shermeyer2018TheImagery} successfully combine super-resolution of satellite imagery with SSD \citep{Liu_2016} and YOLO \citep{redmon2016look} classification networks and report a 13-36\% improvement in mean average precision (mAP) for detecting vehicles in their scenes. They however use much coarser resolution imagery than we have available, changing the nature of the problem, and their proposed approach is based on the VDSR network which has been superseded in more recent work. Extensions of this paper \citep{Ferdous2019SuperDetection} apply \gls{GAN} models, improving vehicle classification accuracy by super-resolving the input imagery. 

In this work we extend these ideas to work on aerial image segmentation, with generalisation to dense labels necessitating high detail and reconstruction accuracy at the pixel level. We improve on prior work by training our \gls{GAN} super-resolution generator jointly with the segmentation network to optimize for the downstream task, inspired by the improvements of task-driven, end-to-end super-resolution training \citep{Haris2018Task-DrivenImages}.

\section{Semantically-Aware Super-Resolution}
\label{sec_Method}

\begin{figure}
\begin{center}
   \includegraphics[width=0.9\linewidth]{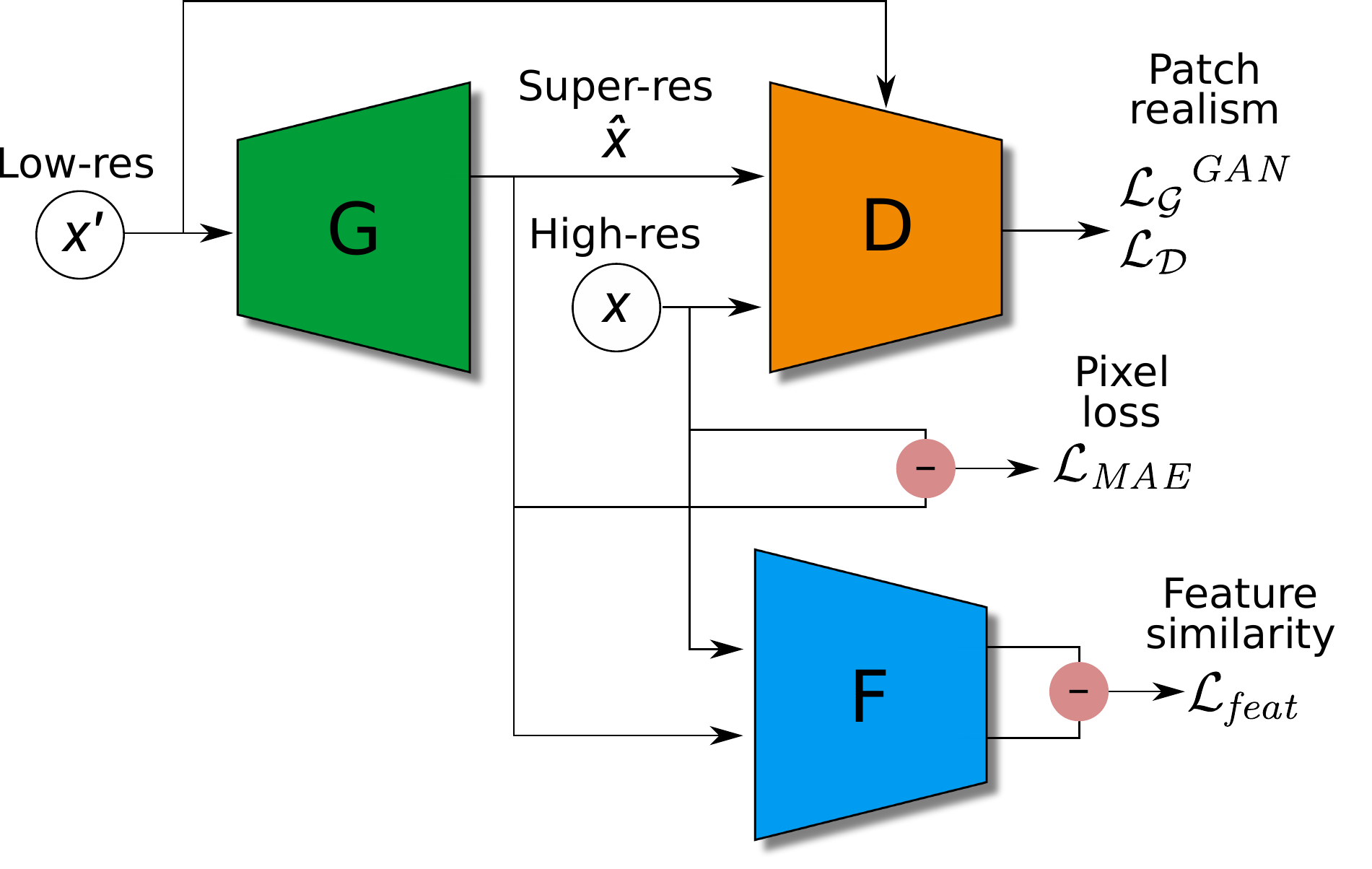}
\end{center}
   \caption{Proposed high-level network architecture: The generator G learns to create super-resolved images $\hat{x}$ from low-resolution images $x'$ based on three losses: the pixel-wise difference between high-resolution and super-resolved signals $\mathcal{L}_{\mathit{MAE}}$; patch realism from the discriminator D conditioned on $x'$; and feature similarity $\mathcal{L}_{\mathit{feat}}$ based on the semantically-driven feature embedding F. 
   }
\label{fig:full_arch}
\end{figure}


To train a super-resolution network we require pairs of high-resolution images $x$ and their corresponding low-resolution counterparts $x'$. For this we begin with full-resolution images drawn from a practical aerial imagery dataset, and construct from each a low-resolution counterpart using a bilinear downsampling operation with a Gaussian anti-aliasing filter. 

Referring to Fig.~\ref{fig:full_arch}, from these image pairs we learn a function $G(x')$ approximating the inverse transform from low to high-resolution image space. We then compute a loss based on pixel-wise comparison of the full-resolution and estimated images, $x$ and $\hat{x}$, as well as feedback from a discriminator D and the quality of the resulting semantic feature mask F.  The choice of loss functions is key to this work and will be covered in depth in Sec.~\ref{section:loss}.

\subsection{Proposed Network Architecture}

Our core network architecture, as illustrated in Fig.~\ref{fig:full_arch}, is best understood as being composed of 3 networks: a generator, a discriminator, as per a typical \gls{GAN} formulation, and a pre-trained semantic segmentation network.

The generator is based on the Residual-in-residual Dense Network (RRDN) architecture proposed in ESRGAN \citep{Wang2019ESRGAN:Networks} as we found this offers superior performance over SRResNet and EDSR models. We however opt for a smaller version with 3 dense blocks rather than the original 16 due to the large number of parameters and thus slow training speeds.

\begin{figure*}
\begin{center}
   \includegraphics[width=\linewidth]{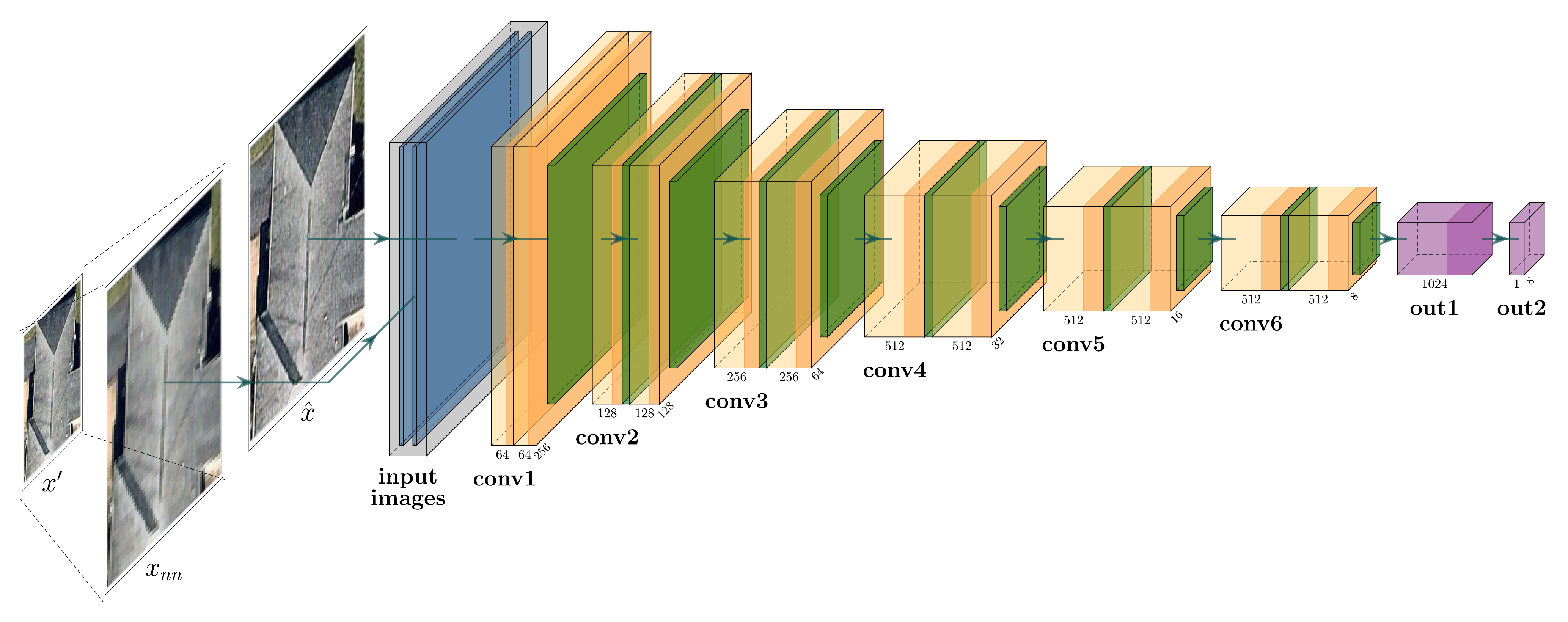}
\end{center}
   \caption{The proposed conditional patch-based discriminator network architecture. Here \texttt{conv} blocks indicate two sets of stacked 3x3 Convolution (yellow) and BatchNorm (green) layers whereas \texttt{out} blocks are single 1x1 Convolutions (purple). Blocks with shaded right hand sides indicate a LeakyReLU activation layer. The low resolution image $x'$ is nearest-neighbour upsampled to produce $x_{nn}$ and then concatenated to the generator prediction $\hat{x}$.}
\label{fig:specific_network_arch}
\end{figure*}

For the discriminator (illustrated in Fig.~\ref{fig:specific_network_arch}) we deviate from a conventional \gls{GAN} image discriminator. We take inspiration from Pix2Pix \citep{Isola2017Image-to-imageNetworks} and employ a patch-based discriminator which outputs predictions over individual ``patches" of the input image. This has been shown to outperform standard discriminator formulations which simply output a binary ``realness" prediction over the entire image. We find that it also allows the network to be fully convolutional and thus compatible with arbitrarily sized input images. Furthermore we apply recent work in Spectral Normalisation \citep{Miyato2018SpectralNetworks} to scale the weights of the discriminator and guarantee Lipschitz-continuity. We find this to be very important in improving the training stability of \glspl{GAN} since having finitely bounded gradients in the discriminator means that for any generated image, the proposed generator avoids spikes in the backpropogation signal and is less likely to destabilise.

We furthermore modify the discriminator to predict realness of images, conditioned on a nearest-neighbour upsample of the low-resolution input image (see Fig.~\ref{fig:specific_network_arch}). This has the benefit of changing the discriminator objective from predicting realistic images standalone to predicting realistic transformation of given source images, which is the true aim of the super-resolution discriminator. This change is also inspired by insight from Isola et al. \citep{Isola2017Image-to-imageNetworks}, demonstrating that conditioning the discriminator on input images significantly improves pixel accuracy and \gls{IoU} scores by penalising generator models with large mismatches between inputs and outputs. This mitigates \gls{GAN} collapse during training in which the generator can fall into a uni-modal state, producing identical outputs regardless of inputs. By maintaining relevance between inputs and outputs, the proposed model can effectively overcome the issue of excessive and unwarranted hallucination of unrealistic detail at higher upsampling factors as is observed in precedent super-resolution works in remote sensing \citep{Bosch2017Super-ResolutionLearning}.

Finally, we use a domain-specific image segmentation network pre-trained to predict semantic masks over aerial imagery for classes such as cars, roads, vegetation, roofs, solar panels and so on. This network is used in parallel with the discriminator to judge image loss, however it focuses on the reconstruction of semantic image features by comparing the semantic segmentation of the original image to the super-resolved prediction. For this we use the segmentation predictions before thresholding for a probabilistic interpretation. By doing so, we naturally imbue our super-resolution network with specific knowledge of the domain learned separately by the segmentation network. This allows the proposed approach to reconstruct image features that are semantically correct with respect to the visual task at hand, in this case image segmentation, as well as optimizing for visual fidelity.

\subsection{Loss Functions} \label{section:loss}

A simple formulation of the loss between two images is the mean across the image of pixel-wise differences, such as the \gls{MSE} or \gls{MAE}. \gls{MAE} is generally regarded to outperform MSE for super-resolution \citep{Johnson2016PerceptualSuper-Resolution} in that it tends to yield sharper image predictions. We have confirmed these findings and thus adopt the \gls{MAE} as our pixel loss function:
\begin{equation}
    \mathcal{L}_{\mathit{MAE}}=\frac{1}{HWC}\left\|x-\hat{x}\right\|_1,
    \label{eq:MAE}
\end{equation}
where $H$, $W$, $C$ are the height, width and channel dimensions of the image respectively and $\hat{x}=G(x')$ is the generator prediction based on our downsampled input image $x'$.

The prominent shortcoming of pixel-wise loss functions is that they compute the average loss over all pixels. An unfortunate consequence of this is that models invariably learn to reconstruct images by trying to minimise their average error and hence produce blurry features that are more correct on average. This effectively masks out small-scale variation and hence fine detail in generated images, which is at odds with the goal of super-resolution. Furthermore, a shortfall of pixel-wise losses is that all losses are calculated at the pixel level and hence have no awareness of macroscopic image composition. Thus, we employ two additional loss terms to remedy these issues, the first being a semantic feature loss and the second being a \gls{GAN} loss. 

Feature loss, referred to as perceptual loss by \cite{Johnson2016PerceptualSuper-Resolution}, was proposed to incorporate the reconstruction of semantic image features by adding a loss term derived from the intermediate activations of an auxiliary classifier network. By feeding both images through this network, the intuition is that if both images have similar semantic content then the classifier activations should be similar. We extend this idea to semantic segmentation networks which produce a spatial mask rather than a single classifier value:
\begin{equation} 
    \mathcal{L}_{\mathit{feat}}=
    \frac{1}{H W C}
    \left\|F(x)-F(\hat{x})\right\|_{2}^{2}%
    ,
    \label{eq:content_loss}
\end{equation}
where $F(x)$ is the embedding of $x$ generated by Nearmap's existing segmentation model pre-trained on high-resolution overhead aerial imagery. We apply the proposed feature loss to train a model to reconstruct important semantic features such as roofing material, roads, etc. Rather than using scalar classification predictions, we calculate the binary cross-entropy loss across predictions on patches of the image, similar to Pix2Pix \citep{Isola2017Image-to-imageNetworks}.

Finally, we introduce a \gls{GAN} loss term following the Non-Saturating \gls{GAN} (NS-GAN) formulation \citep{goodfellow2014generative},
\begin{equation}
    \mathcal{L_G}^{\mathit{GAN}} = - \mathbb{E}_{\hat{x}\sim p_g}[\log(D(\hat{x}))]  \label{eq:g_loss},
\end{equation}
where $D(x)$ is the discriminator's estimate of patch realism.
We alter the discriminator loss to be conditional on the low-resolution input image, 
\begin{equation}
\begin{aligned} \label{eq:d_loss}
    \mathcal{L_D} = &-\mathbb{E}_{x\sim p_d}[\log(D(x|x_{\mathit{nn}}))] \\
    &- \mathbb{E}_{\hat{x}\sim p_g}[\log(1-D(\hat{x}|x_{\mathit{nn}}))],
\end{aligned}
\end{equation}
so the discriminator learns not just whether a given high-resolution image is ``realistic", but rather whether it is a realistic super-resolution of a given low-resolution image. We implement this with the nearest-neighbour upsampling of the low-resolution image, $x_{nn}$, similar to learning the residual \citep{He2016DeepRecognition}.

As is common in combining loss functions~\citep{shen2020frustum}, we employ weighting hyperparameters
\begin{equation}
    \mathcal{L_G} = \alpha\mathcal{L_G}^{\mathit{GAN}} + \beta\mathcal{L}_{\mathit{feat}} + \gamma\mathcal{L}_{\mathit{MAE}},
    \label{eq:overall_loss}
\end{equation}
empirically selecting $\alpha = 1\times 10^{-3}$, $\beta = 5$ and $\gamma = 1\times 10^{-3}$ to achieve a balance between imagined detail and true semantic reconstruction.

\section{Results}
\label{sec_Results}

\subsection{Dataset and Training}

\begin{figure}
\begin{center}
   \includegraphics[width=1.0\linewidth]{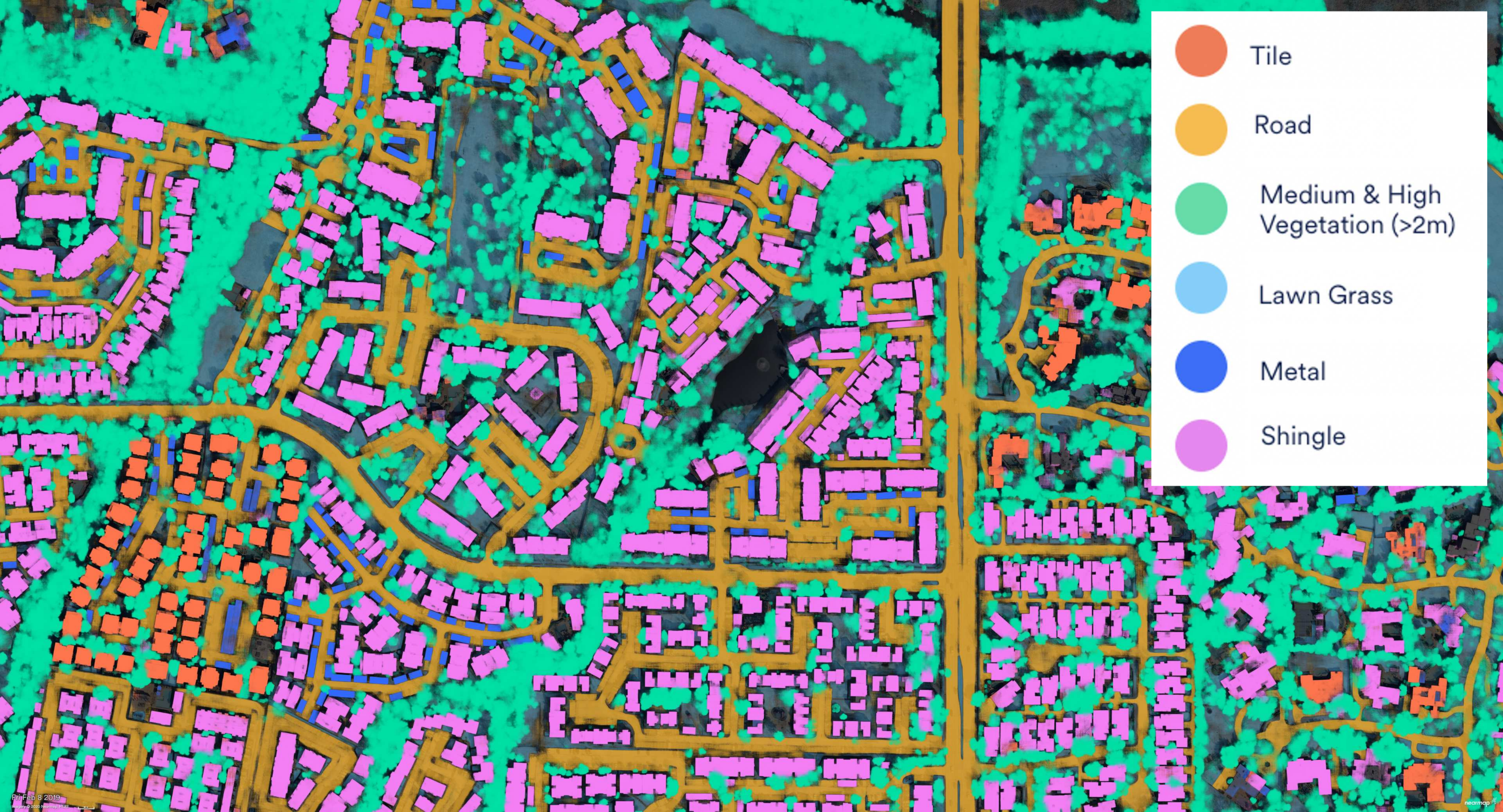}
\end{center}
   \caption{Typical raster image from dataset, illustrating the predicted segmentation masks for the many landmarks and materials of interest within the scene. For more information visit the Nearmap AI Documentation \url{https://docs.nearmap.com/display/ND/NEARMAP+AI}.}
\label{fig:nearmap_data}
\end{figure}

For training we use a very large-scale dataset of nearly half a million aerial images accompanied by human annotated feature masks, an example of which is portrayed in Fig.~\ref{fig:nearmap_data}. The dataset covers imagery across Australia and the United States, captured at a \gls{GSD} resolution of 5-7~cm. In this work, we used labels for 34 unique categories of typical aerial image features including specific object instances (swimming pool, solar panel), materials (shingle roof, tile roof), and other common features (water body, shadow).

We trained our super-resolution models using Tensorflow 2 on an NVIDIA V100 GPU, using a batch size of 16. We employed a two-stage training procedure of first performing supervised pre-training of the generator with an \gls{MAE} pixel-wise loss (Eq.~\ref{eq:MAE}), then fine-tuned with the discriminator network in an adversarial scenario. This approach significantly accelerated convergence and helped avoid instability characteristic of \gls{GAN} models in early iterations of training. 

We employed the ADAM optimiser \citep{Kingma2014Adam:Optimization}, using a pre-training learning rate of $lr_g=2\times 10^{-4}$. For fine-tuning we employed different initial learning rates for the generator and discriminator networks, starting at $lr_g=1\times 10^{-4}$ and $lr_d=2\times 10^{-4}$ respectively. We found that decaying these by a factor of one half every 10,000 optimizer steps yielded reliable \gls{GAN} convergence. Typically we trained for between 10,000 and 40,000 iterations with longer training required for the higher upsampling factors. Further training showed a regression in quality associated with over-fitting.

To accelerate training and performance at higher upsampling rates, we initialised the model with weights trained for lower upsampling rates. This is consistent with previous works showing super-resolution models benefit from information learned at multiple scales \citep{Lim2017EnhancedSuper-Resolution}, and leave the exploration of multi-scale architectures for future work.


\subsection{Image Quality Metrics}

To evaluate image quality we employ three well-known full-reference image metrics: 
\Gls{PSNR} measured in dB as  
\begin{equation}
\operatorname{PSNR}=10 \cdot \log _{10}\left(\frac{\operatorname{MAX}_{I}^{2}}{\operatorname{MSE}}\right),
\label{eq:PSNR}
\end{equation}
where $\operatorname{MAX_I}$ is the maximum pixel intensity in the image and $\operatorname{MSE}$ is the mean-squared error. 

\Gls{SSIM}, designed to more closely reflect subjective evaluation of image fidelity:
\begin{equation}
\operatorname{SSIM}(x, y)=\frac{\left(2 \mu_{x} \mu_{y}+c_{1}\right)\left(2 \sigma_{x y}+c_{2}\right)}{\left(\mu_{x}^{2}+\mu_{y}^{2}+c_{1}\right)\left(\sigma_{x}^{2}+\sigma_{y}^{2}+c_{2}\right)},
\label{eq:SSIM}
\end{equation}
where $\mu$ and $\sigma$ are the mean and standard deviation of their respective images and $c1$, $c2$ are constants proportional to the dynamic range of the given images.

And finally \gls{MS-SSIM}, which extends \gls{SSIM} as a weighted average at various downsampled resolutions. The latter is often reported for super-resolution tasks as these deal with judging image quality over several scales. We employed the implementations of these metrics found in Tensorflow's image module\footnote{\url{www.tensorflow.org/api_docs/python/tf/image}}.

\subsection{Image Segmentation Metrics}
One of the most commonly used metrics in semantic segmentation is the \gls{IoU}, also referred to as the Jaccard Index. It compares predicted and known target masks $A$ and $B$ by taking the ratio
\begin{equation}
\operatorname{IoU}(A, B)=\frac{|A \cap B|}{|A \cup B|}.
\label{eq:iou}
\end{equation}
Since we aim to improve segmentation accuracy over a multi-class dataset, we report the class-wise \gls{mIoU}, taken as the mean IoU over label classes \citep{LongFullySegmentation}. 

We also report the percentage improvement of the proposed method $A$ relative to a baseline method $B$ as
\begin{equation}
\operatorname{\% Improvement}(A, B)=\frac{A-B}{B} \cdot 100.
\label{eq:perc_improvement}
\end{equation}


\subsection{Super-Resolution Performance}

\begin{table}
\caption{A comparison of of three potential generator model architectures for 4$\times$ super-resolution. We chose RRDN for all subsequent experiments due to its superior overall performance on pixel-wise loss objectives.} \vspace{-1em}
\begin{center}
\begin{tabular}{ccccc}
\hline
Model             & MAE Loss  & PSNR  & SSIM    & MS-SSIM\\ \hline \hline
SRResNet               & 0.04066    & 25.44      & 0.6919 & 0.9296\\ \hline
EDSR                          & 0.04059    & 25.45      & 0.6930 & 0.9301\\ \hline
RRDN                          & \textbf{0.04035}    & \textbf{25.51}   & \textbf{0.6957} &   \textbf{0.9308} \\ \hline
\end{tabular}
\end{center}
\label{tbl:pixelloss}
\end{table}


To select a generator model architecture, we compared four alternatives for 4$\times$ super-resolution using only a pixel-wise loss objective. Results are summarized in Table~\ref{tbl:pixelloss}, showing that the RRDN-style backbone performed best at this task. 

\begin{figure}
\captionsetup[subfloat]{justification=centering, font=scriptsize}
\centering
\includegraphics[width=0.24\linewidth]{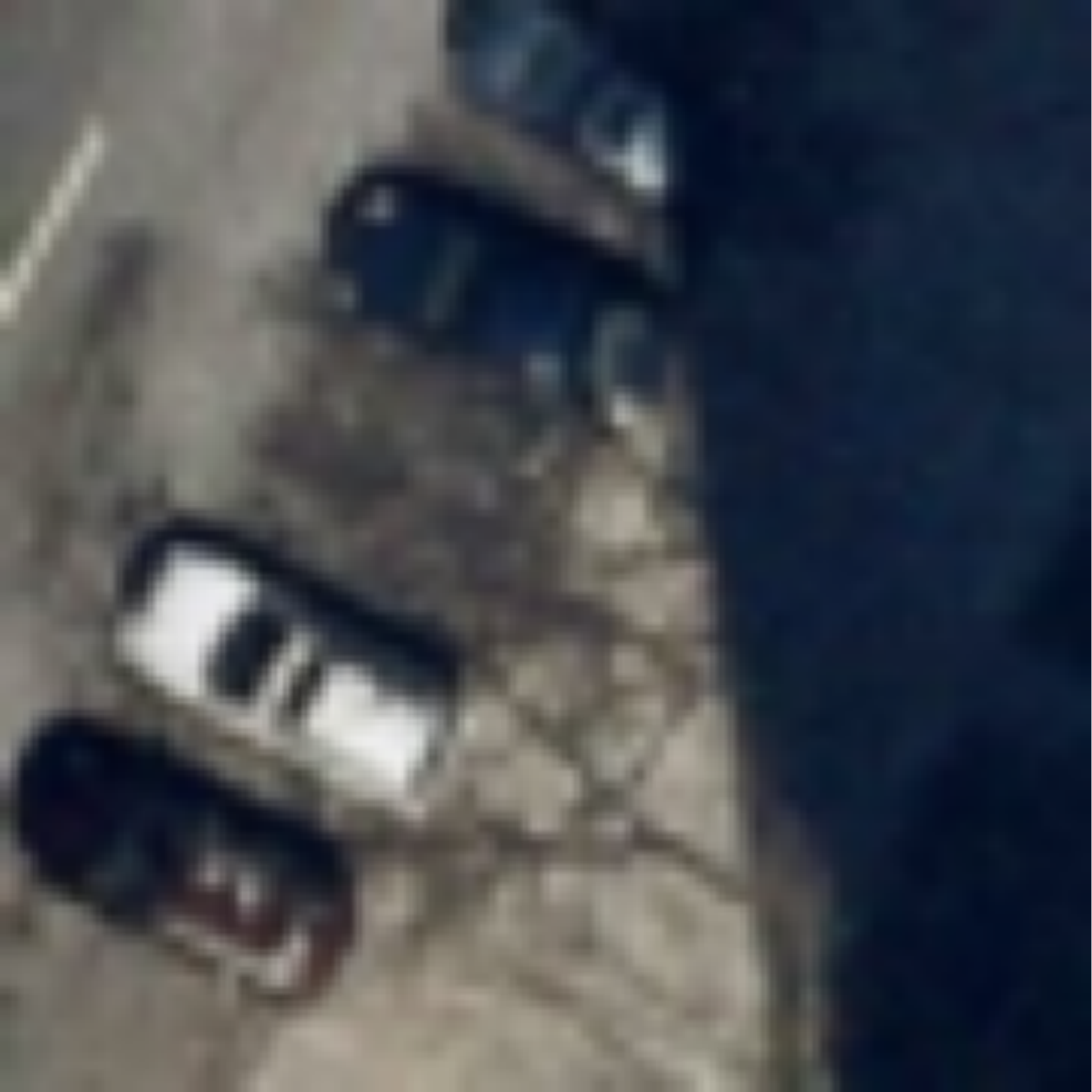}%
\hfill
\includegraphics[width=0.24\linewidth]{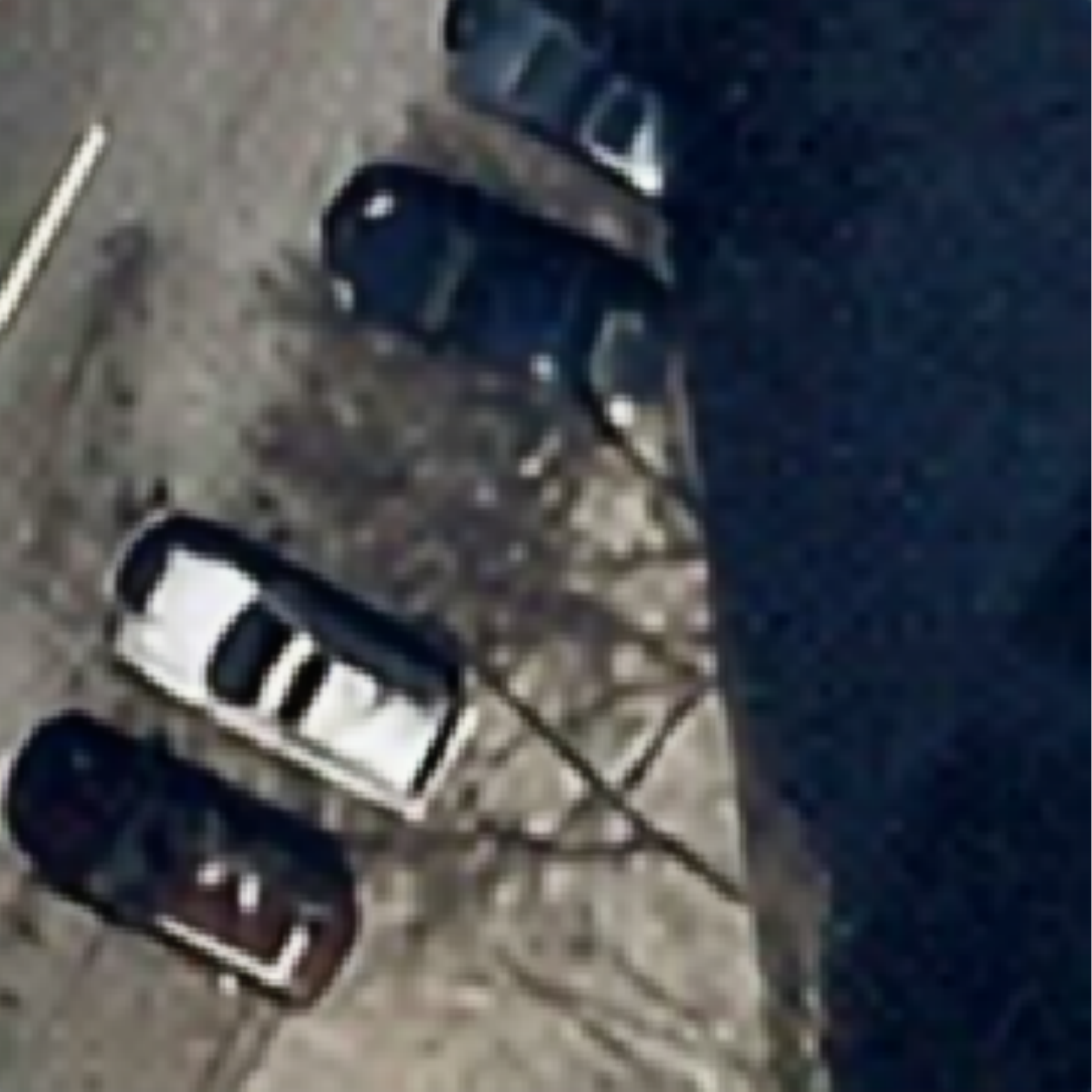}%
\hfill
\includegraphics[width=0.24\linewidth]{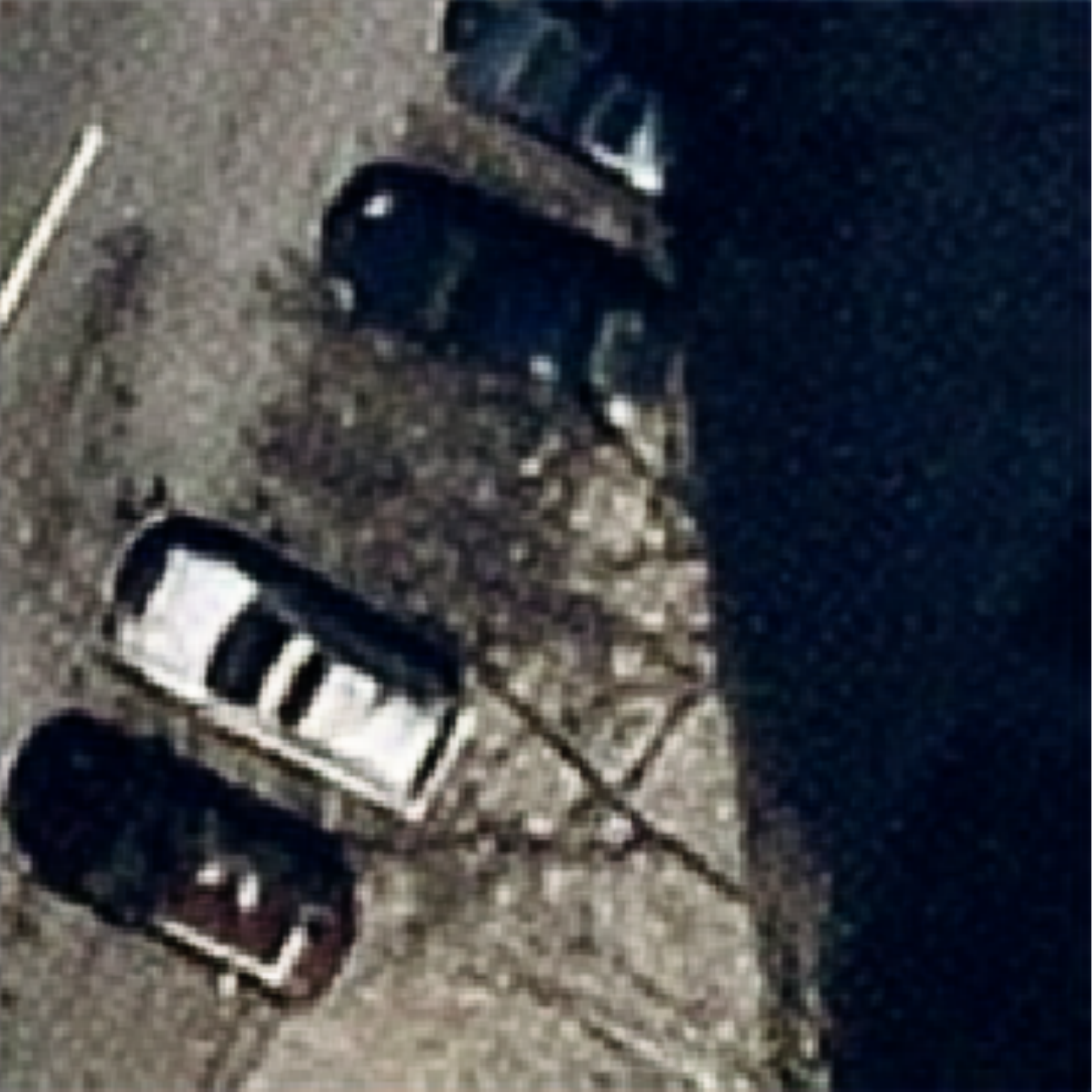}%
\hfill
\includegraphics[width=0.24\linewidth]{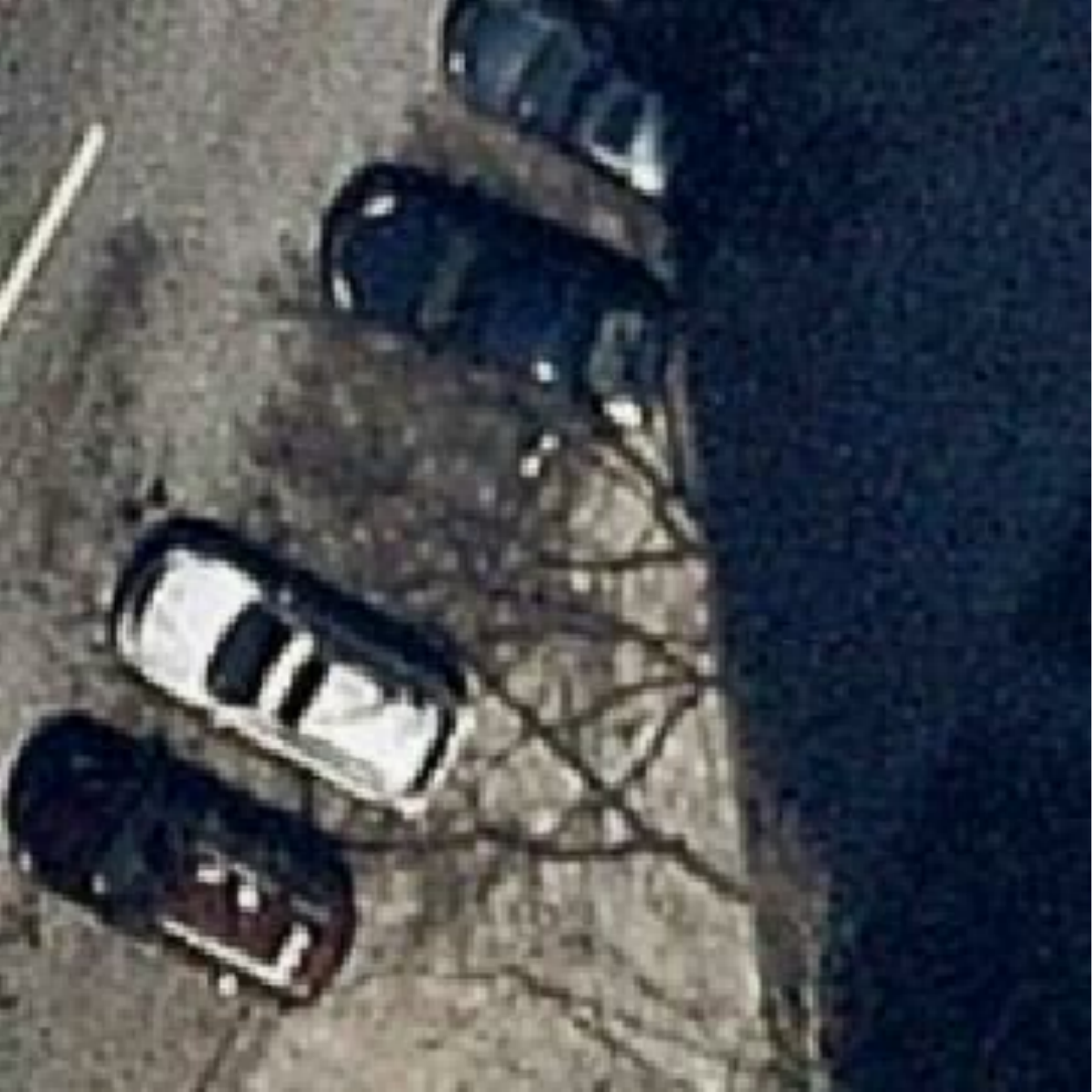}%

\subfloat[][Bilinear (4x)\\ (22.94 / 0.58 / 0.86 / 0.19)]{\includegraphics[width=0.24\linewidth]{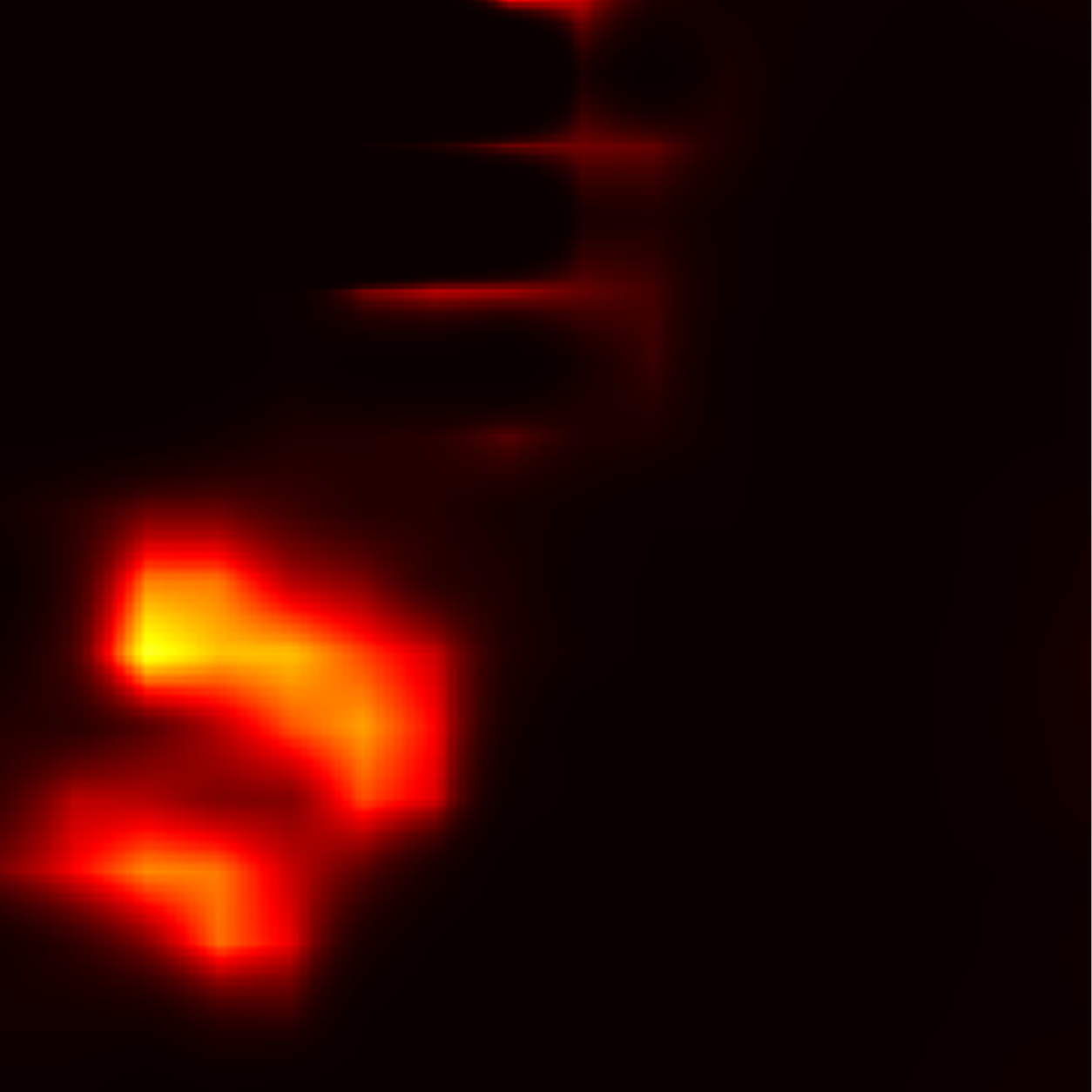}}%
\hfill
\subfloat[][RRDN (4x)\\ (\textbf{27.05} / \textbf{0.74} / \textbf{0.94} / 0.52)]{\includegraphics[width=0.24\linewidth]{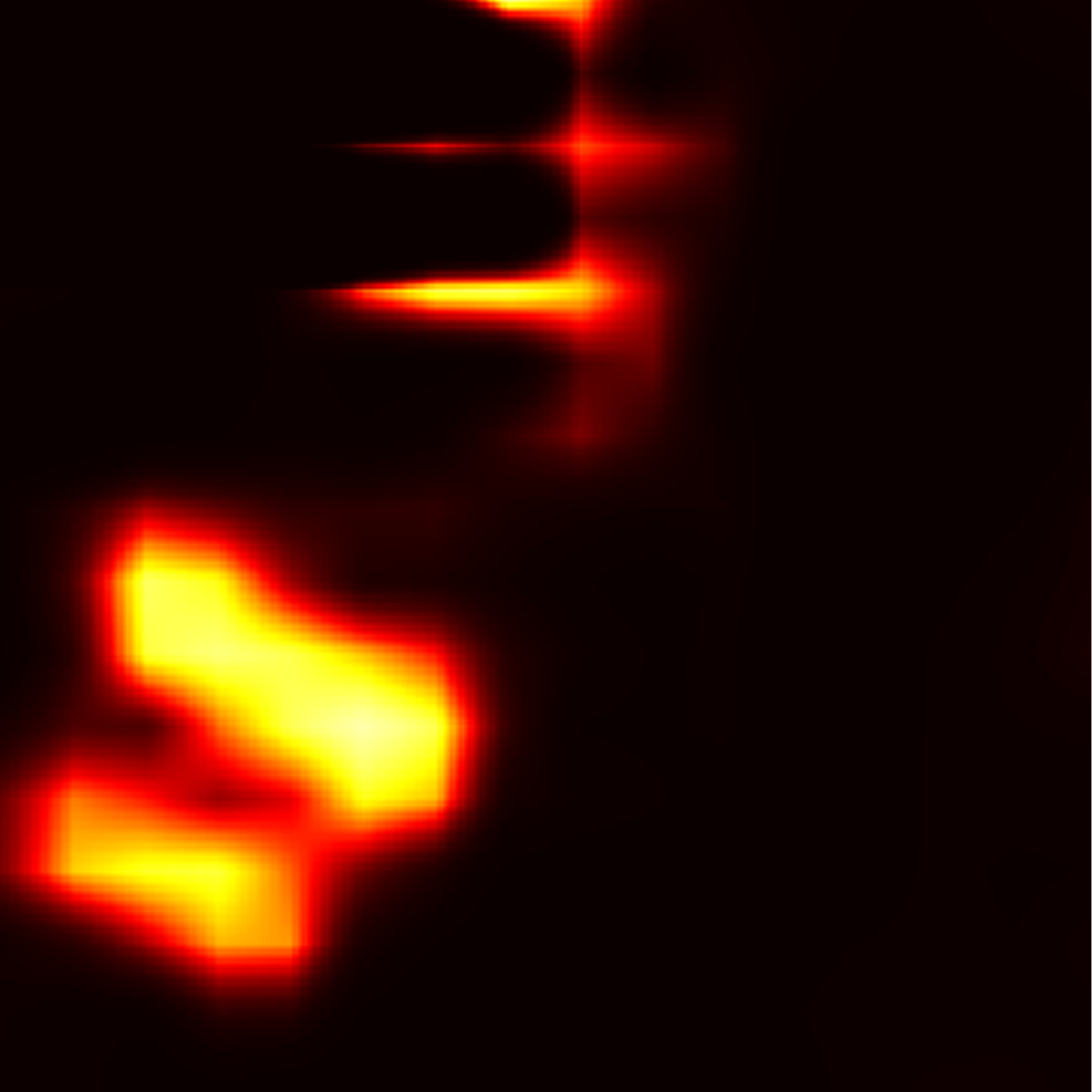}}%
\hfill
\subfloat[][Ours (4x)\\ (23.48 / 0.60 / 0.91 / \textbf{0.77})]{\includegraphics[width=0.24\linewidth]{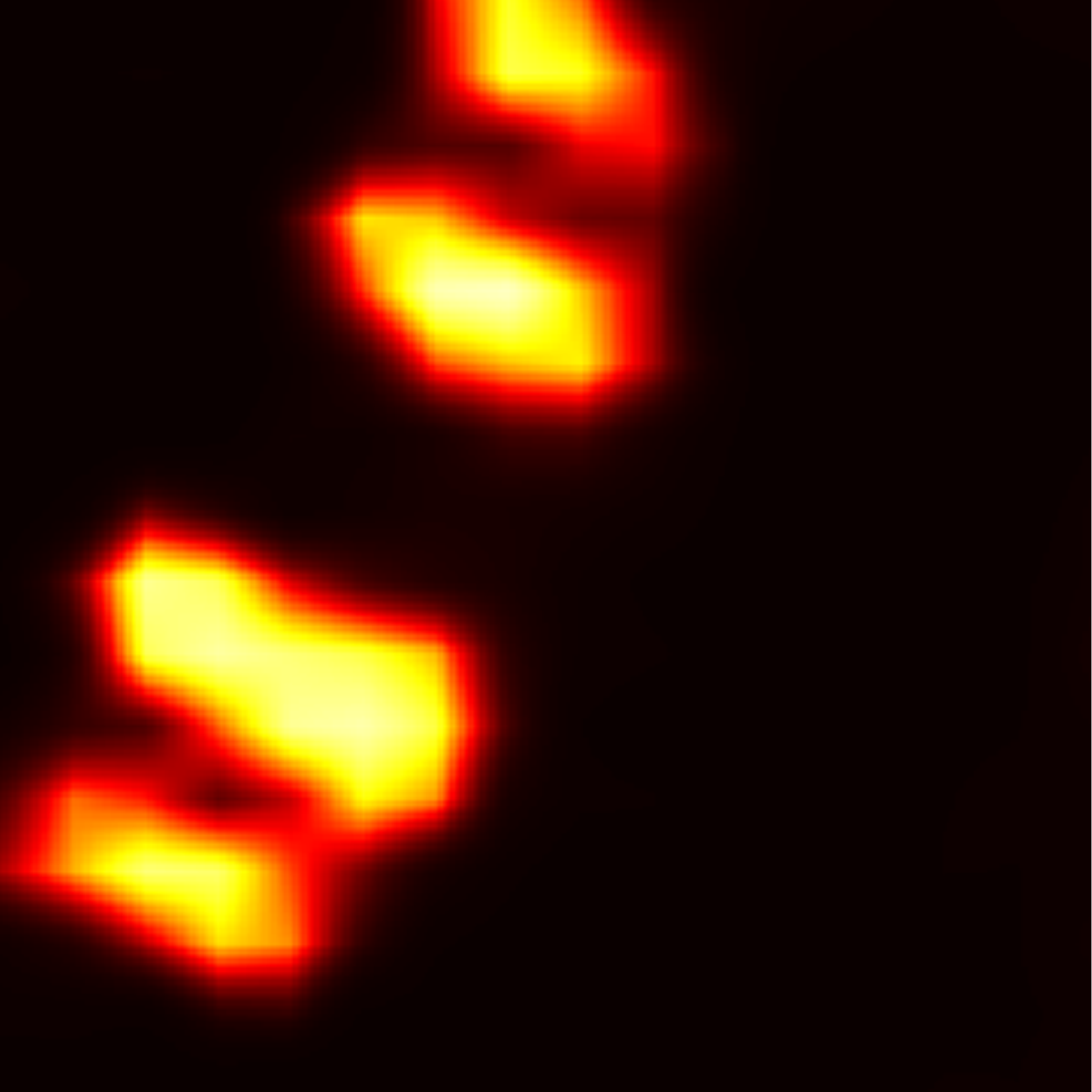}}%
\hfill
\subfloat[][HR (4x)\\ ($\infty$ / 1.00 / 1.00 / 1.00)]{\includegraphics[width=0.24\linewidth]{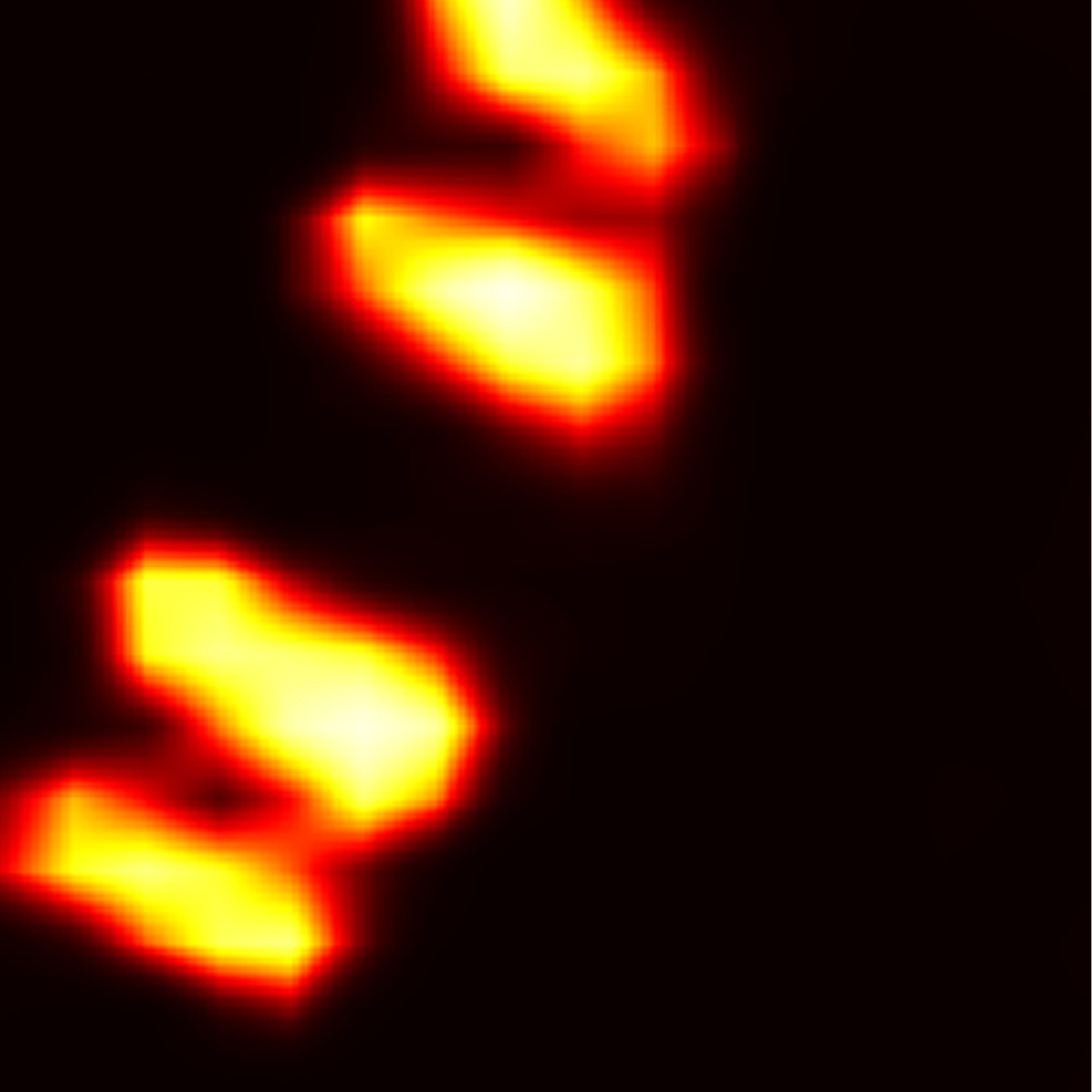}}%

\includegraphics[width=0.24\linewidth]{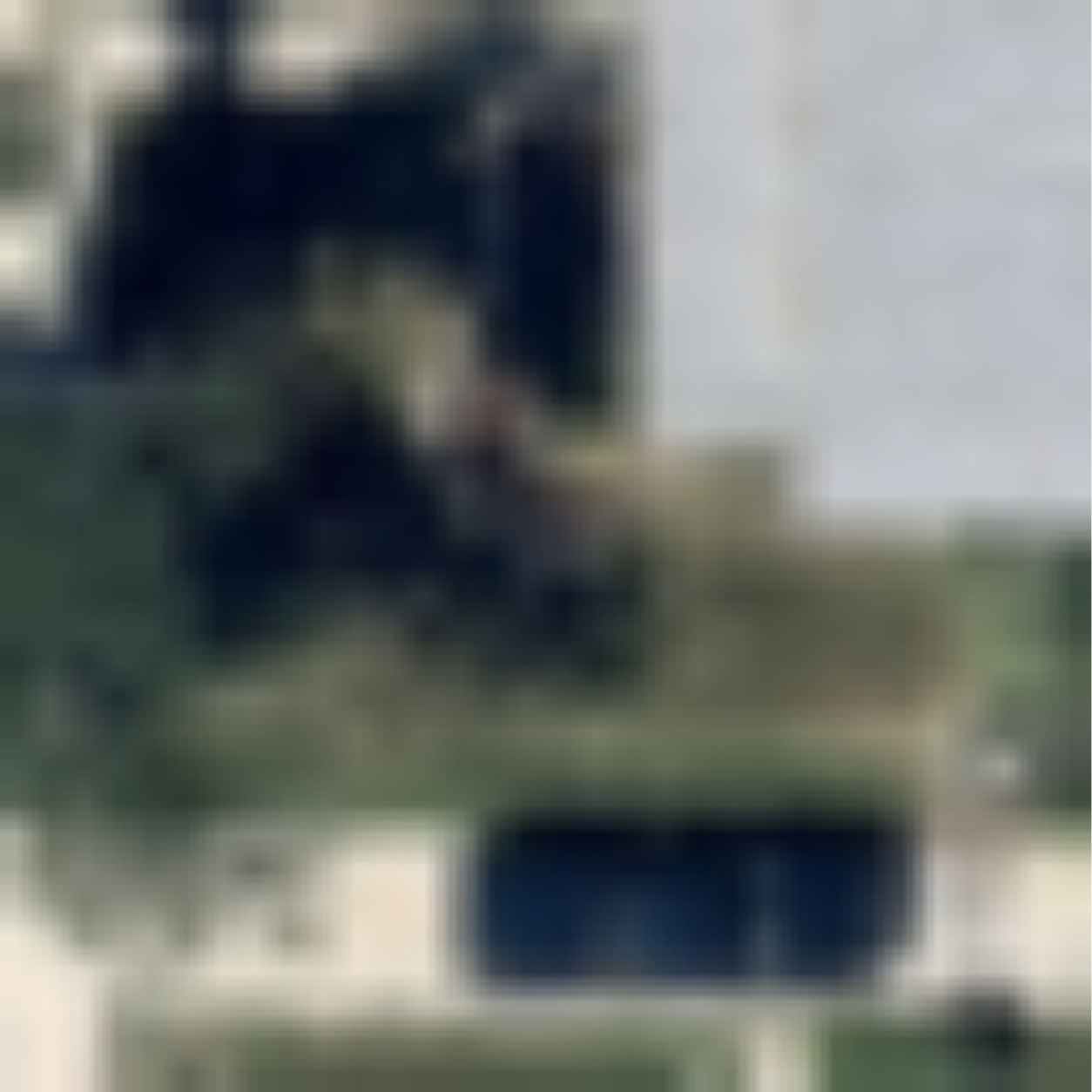}%
\hfill
\includegraphics[width=0.24\linewidth]{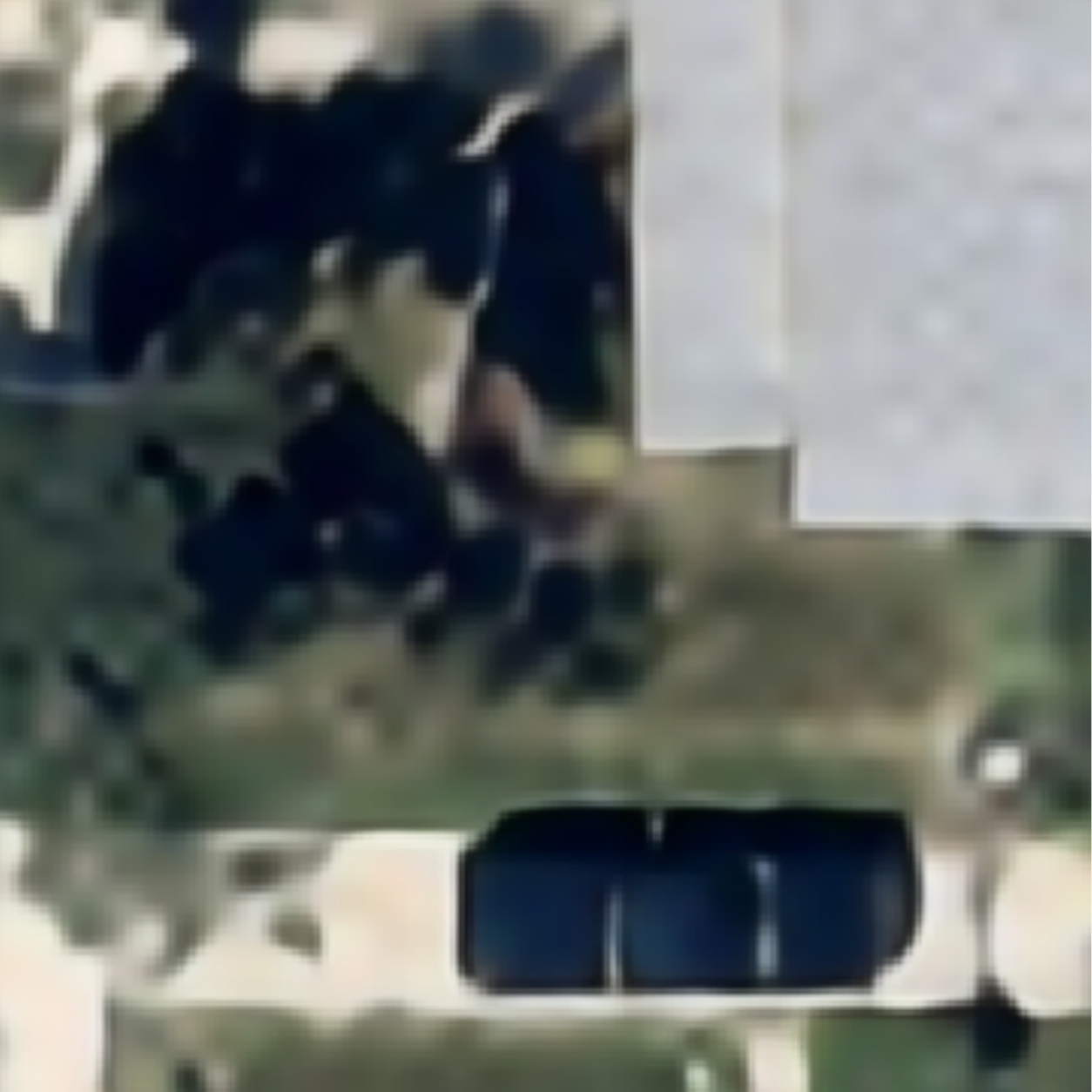}%
\hfill
\includegraphics[width=0.24\linewidth]{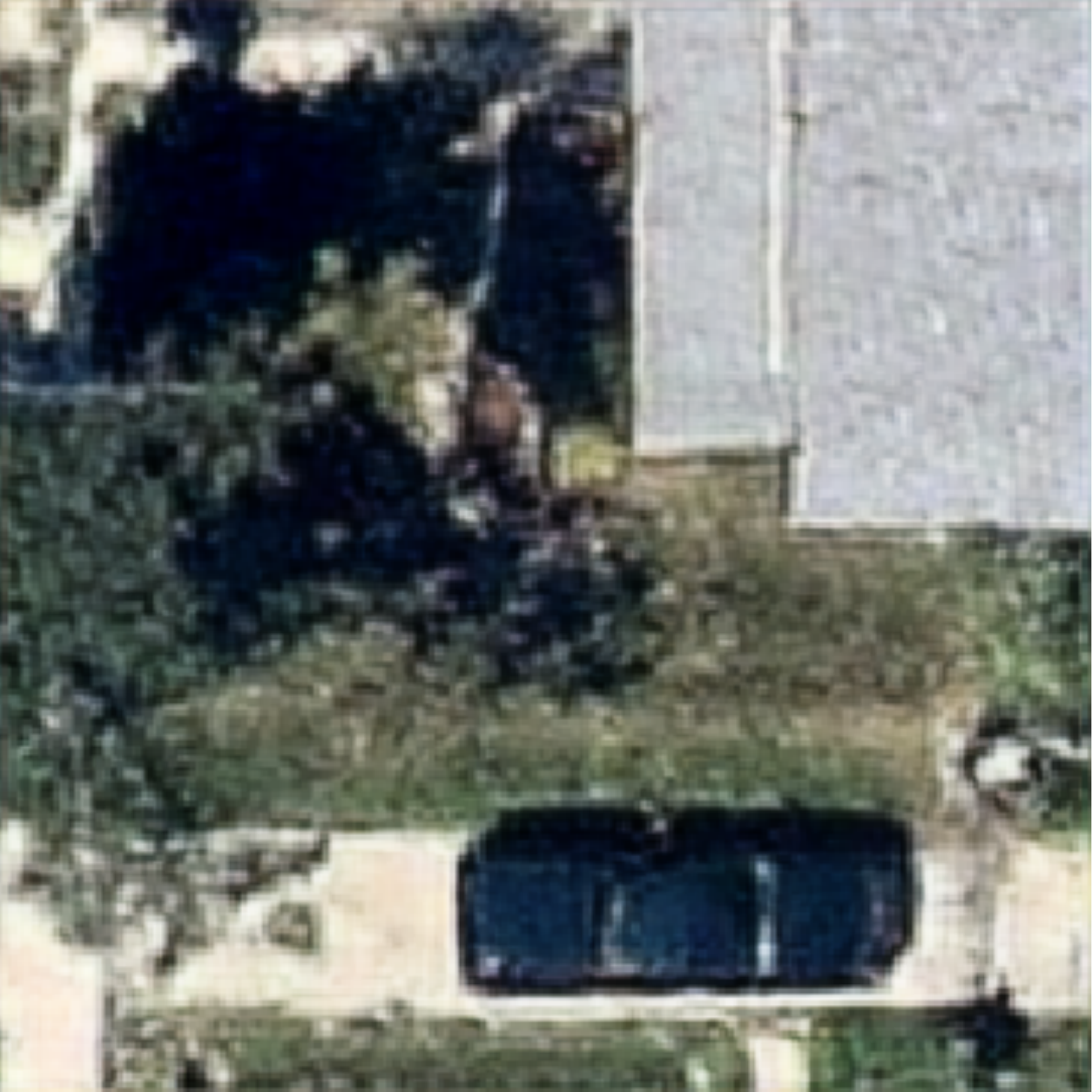}%
\hfill
\includegraphics[width=0.24\linewidth]{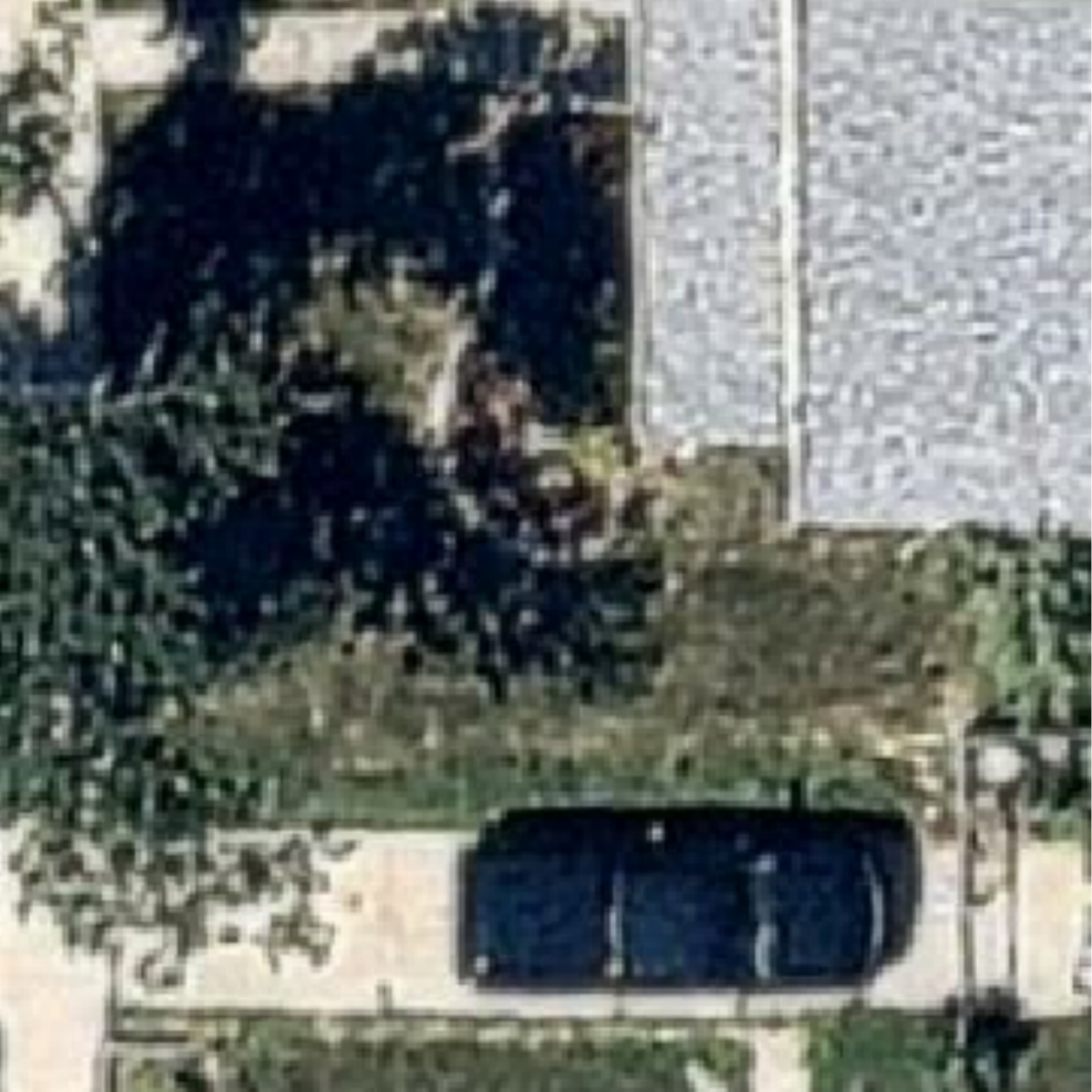}%

\subfloat[][Bilinear (8x)\\ (18.79 / 0.29 / 0.61 / 0.00)]{\includegraphics[width=0.24\linewidth]{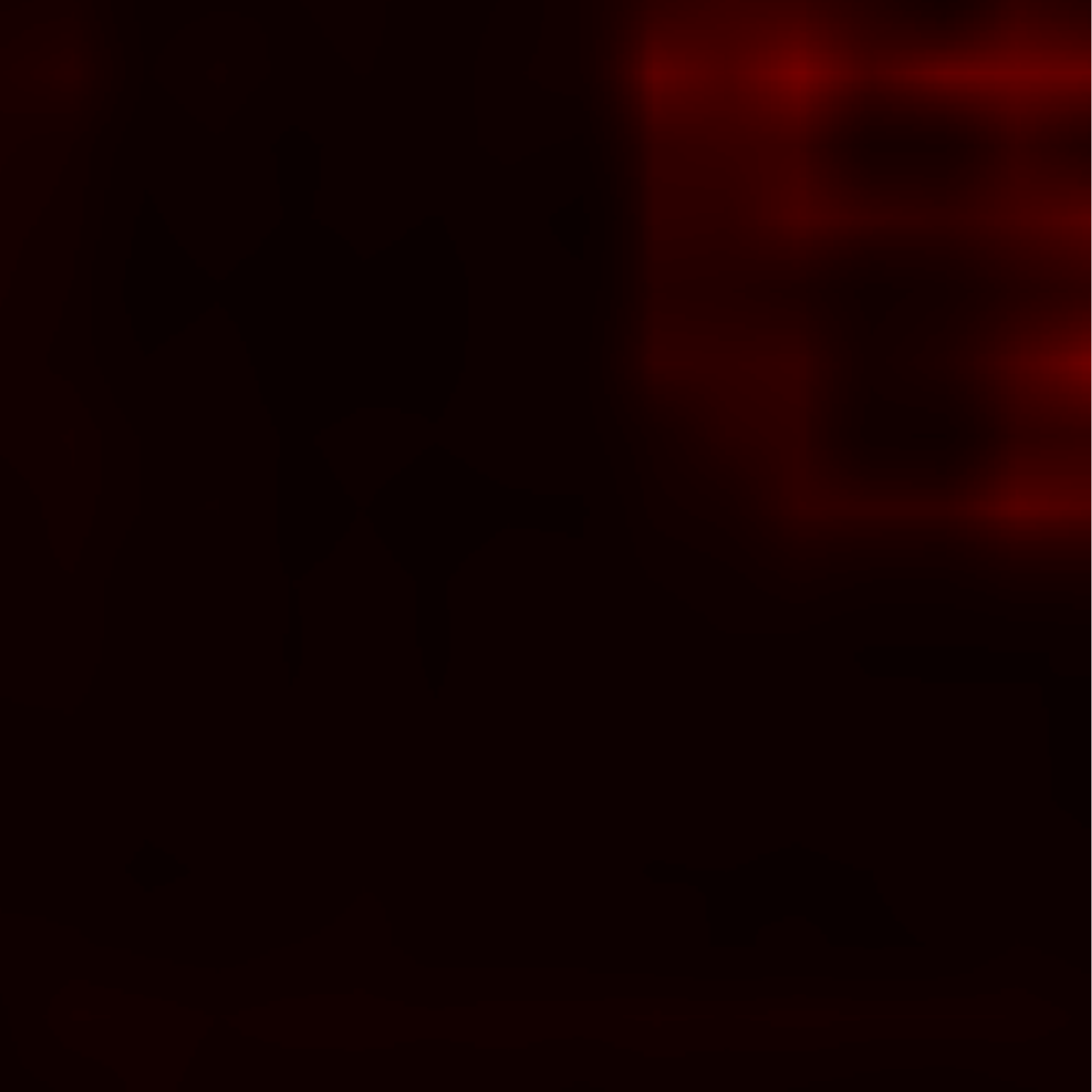}}%
\hfill
\subfloat[][RRDN (8x)\\ (\textbf{20.15} / \textbf{0.39} / \textbf{0.73} / 0.20)]{\includegraphics[width=0.24\linewidth]{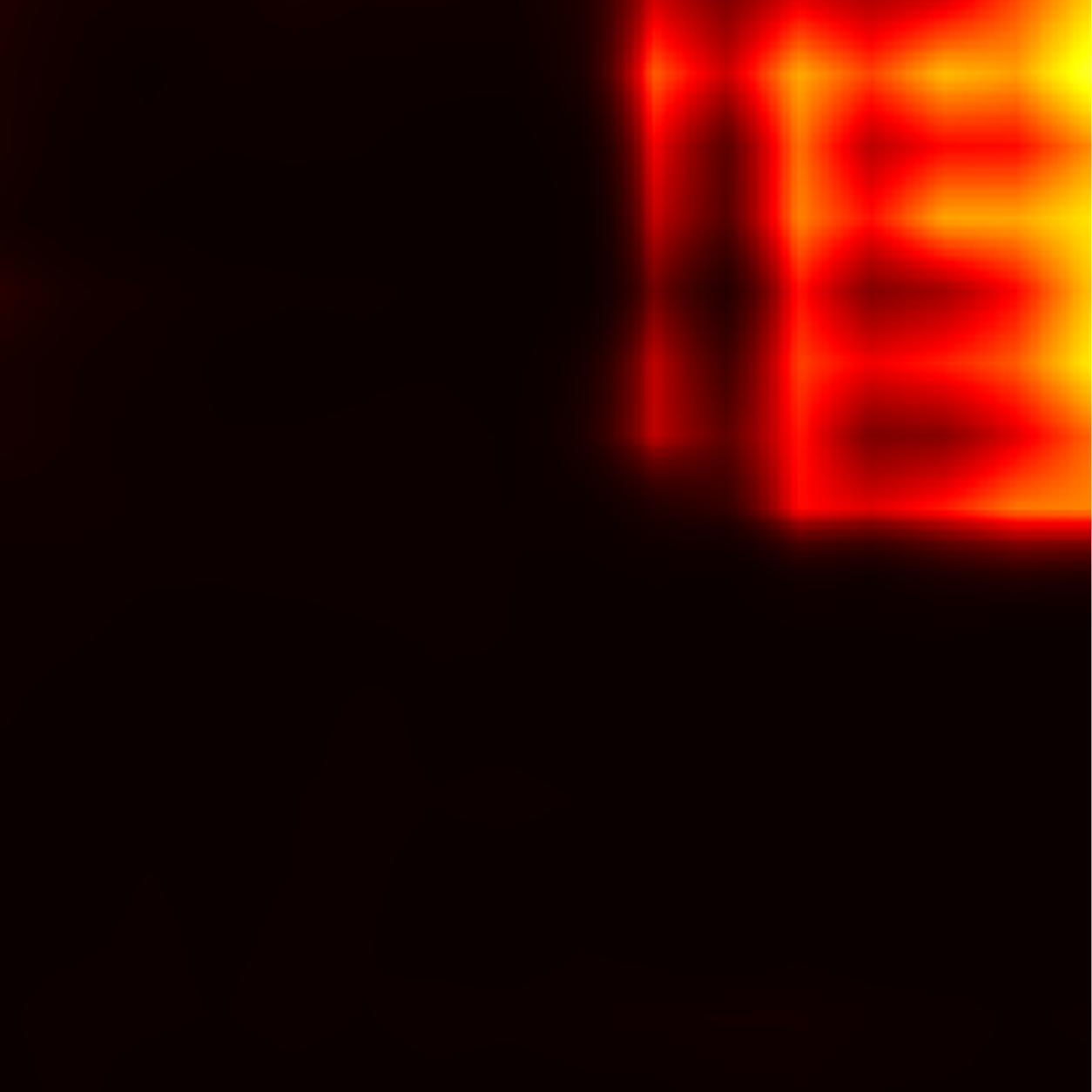}}%
\hfill
\subfloat[][Ours (8x)\\ (18.65 / 0.32 / 0.69 / \textbf{0.97})]{\includegraphics[width=0.24\linewidth]{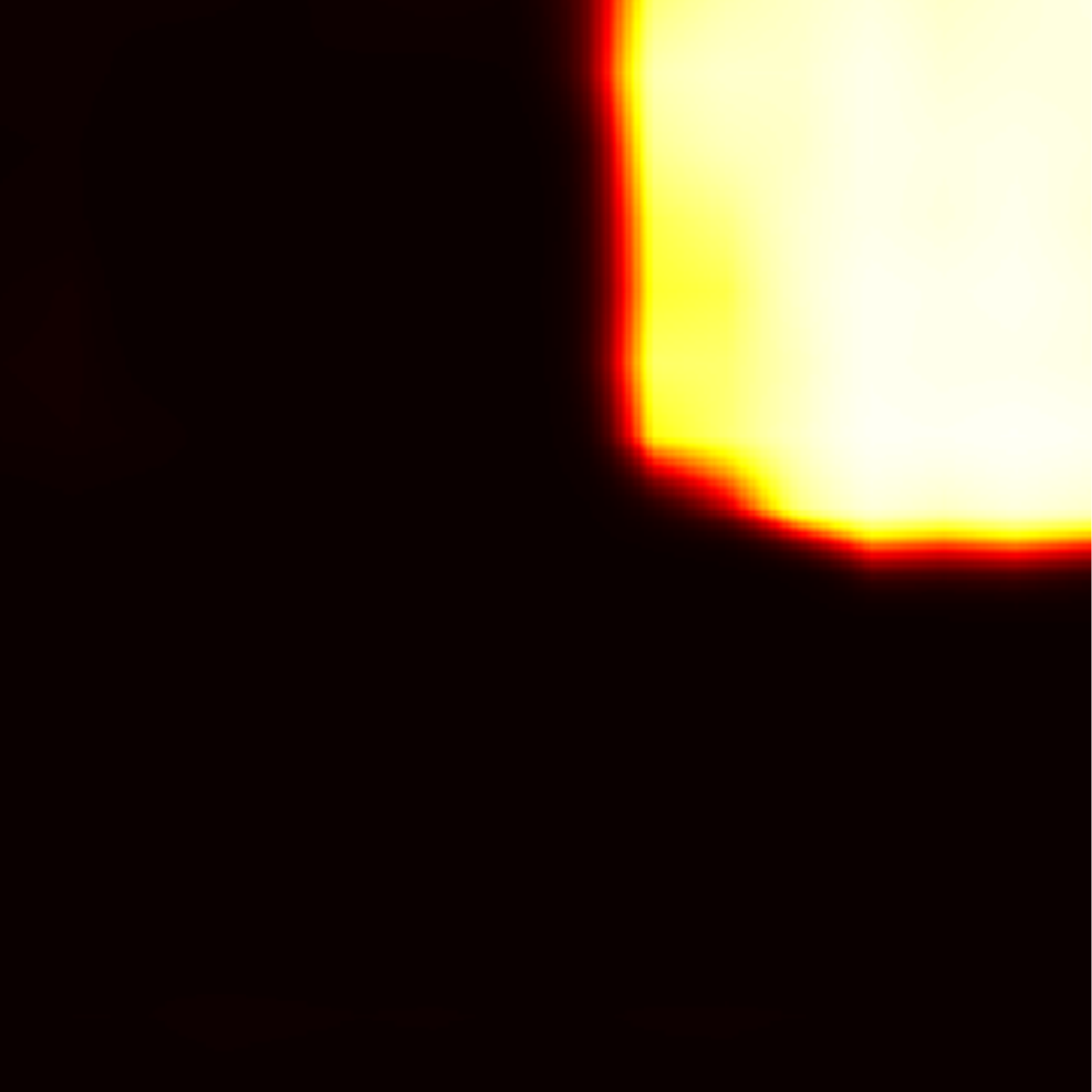}}%
\hfill
\subfloat[][HR (8x)\\ ($\infty$ / 1.00 / 1.00 / 1.00)]{\includegraphics[width=0.24\linewidth]{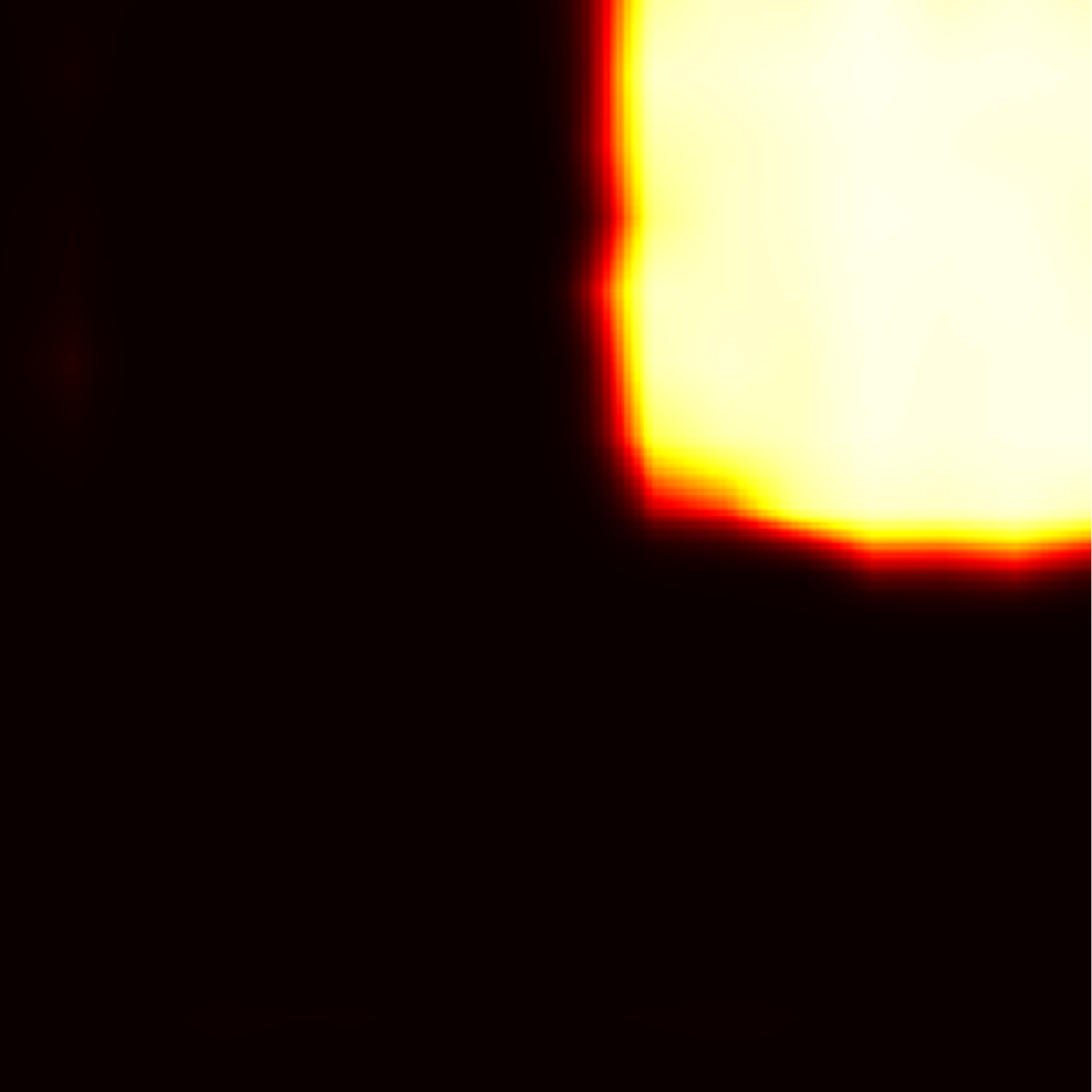}}%

\includegraphics[width=0.24\linewidth]{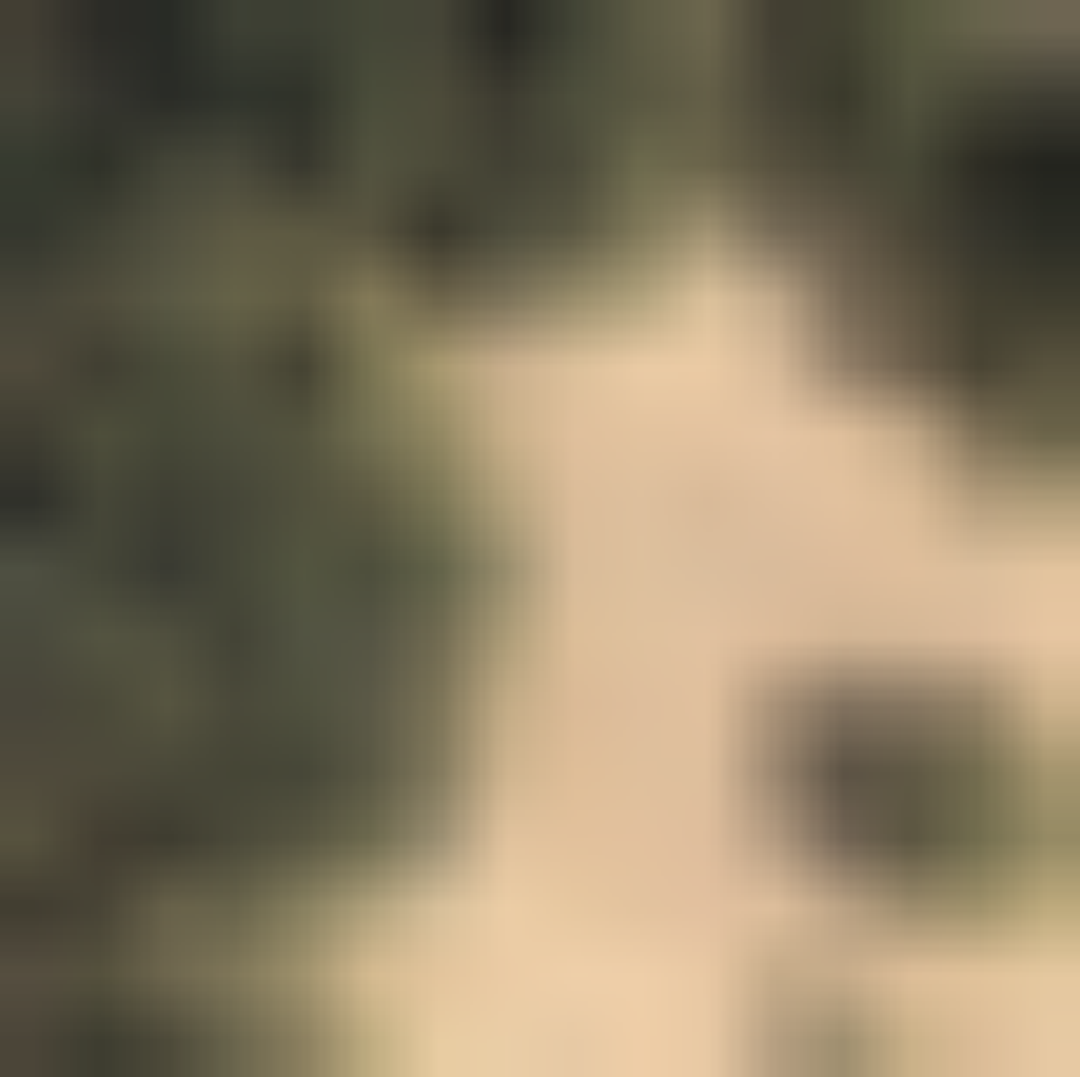}%
\hfill
\includegraphics[width=0.24\linewidth]{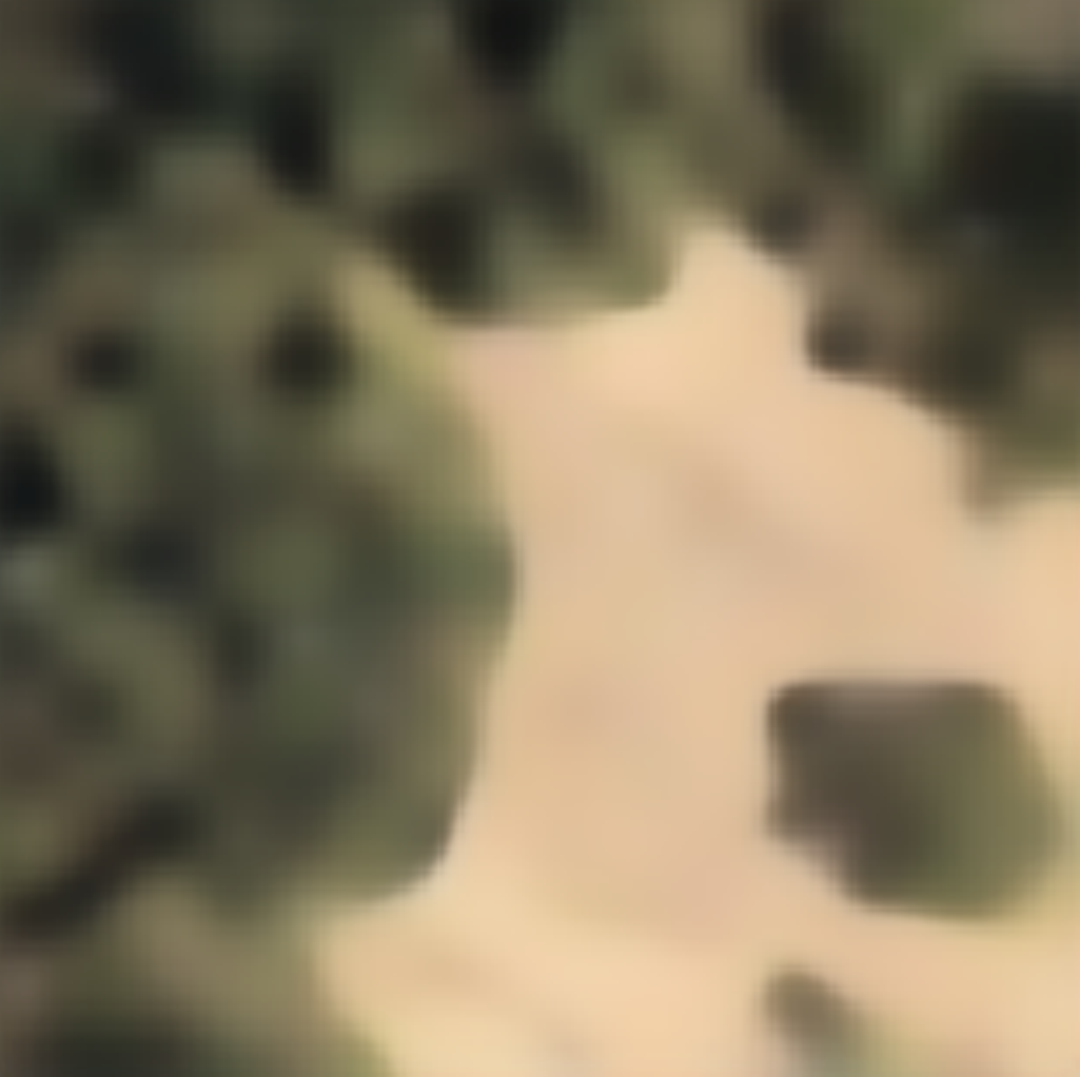}%
\hfill
\includegraphics[width=0.24\linewidth]{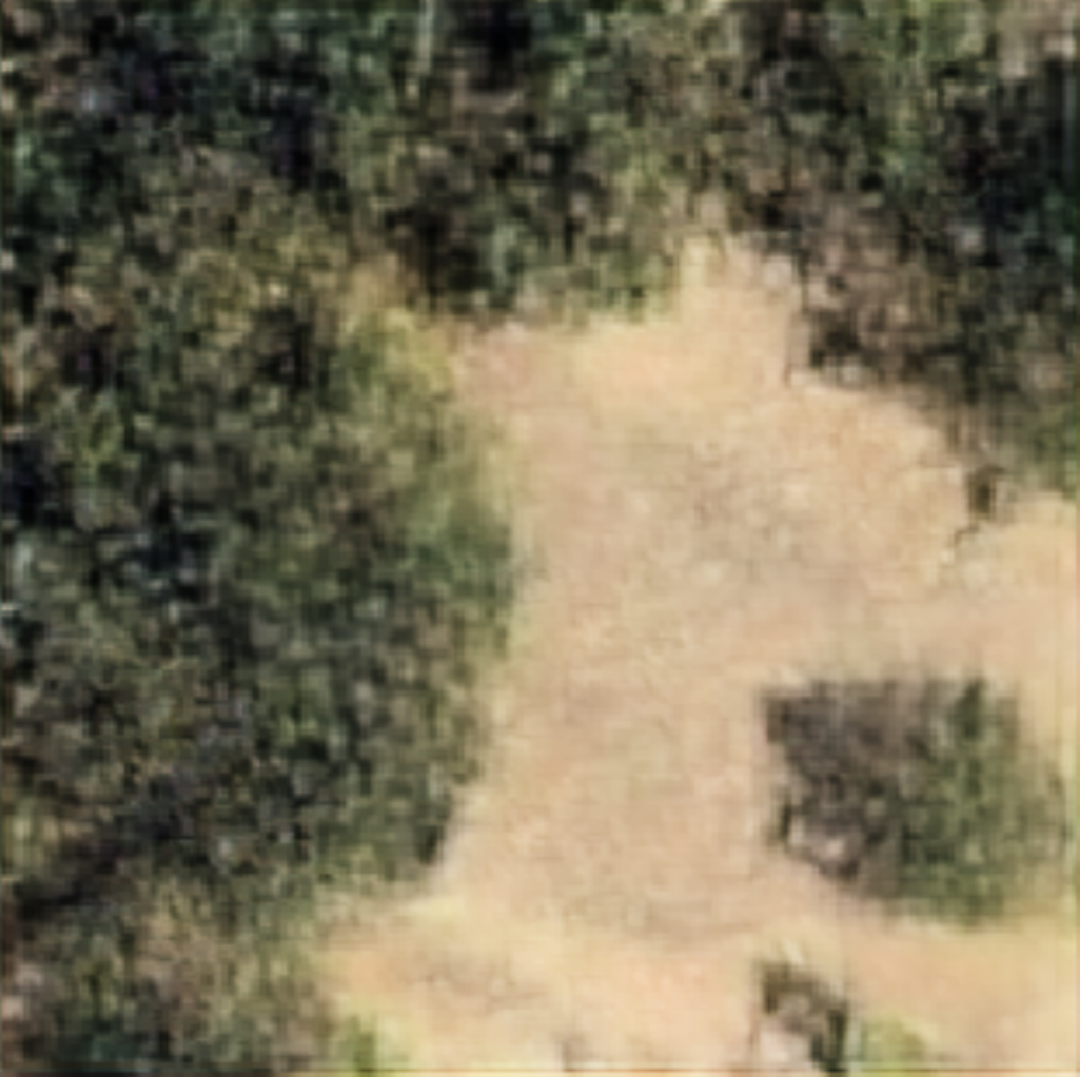}%
\hfill
\includegraphics[width=0.24\linewidth]{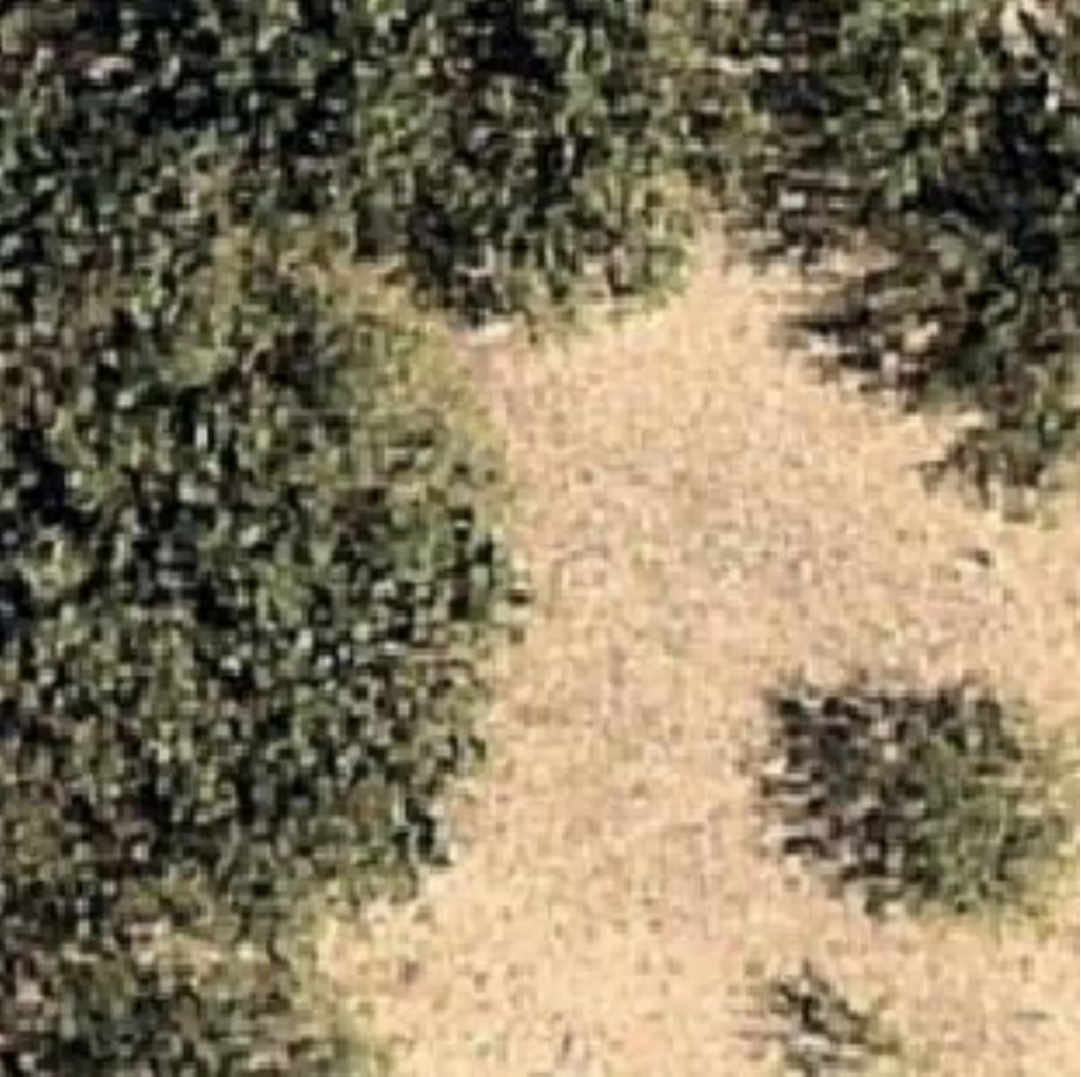}%

\subfloat[][Bilinear (16x)\\ (17.59 / 0.17 / 0.43 / 0.00)]{\includegraphics[width=0.24\linewidth]{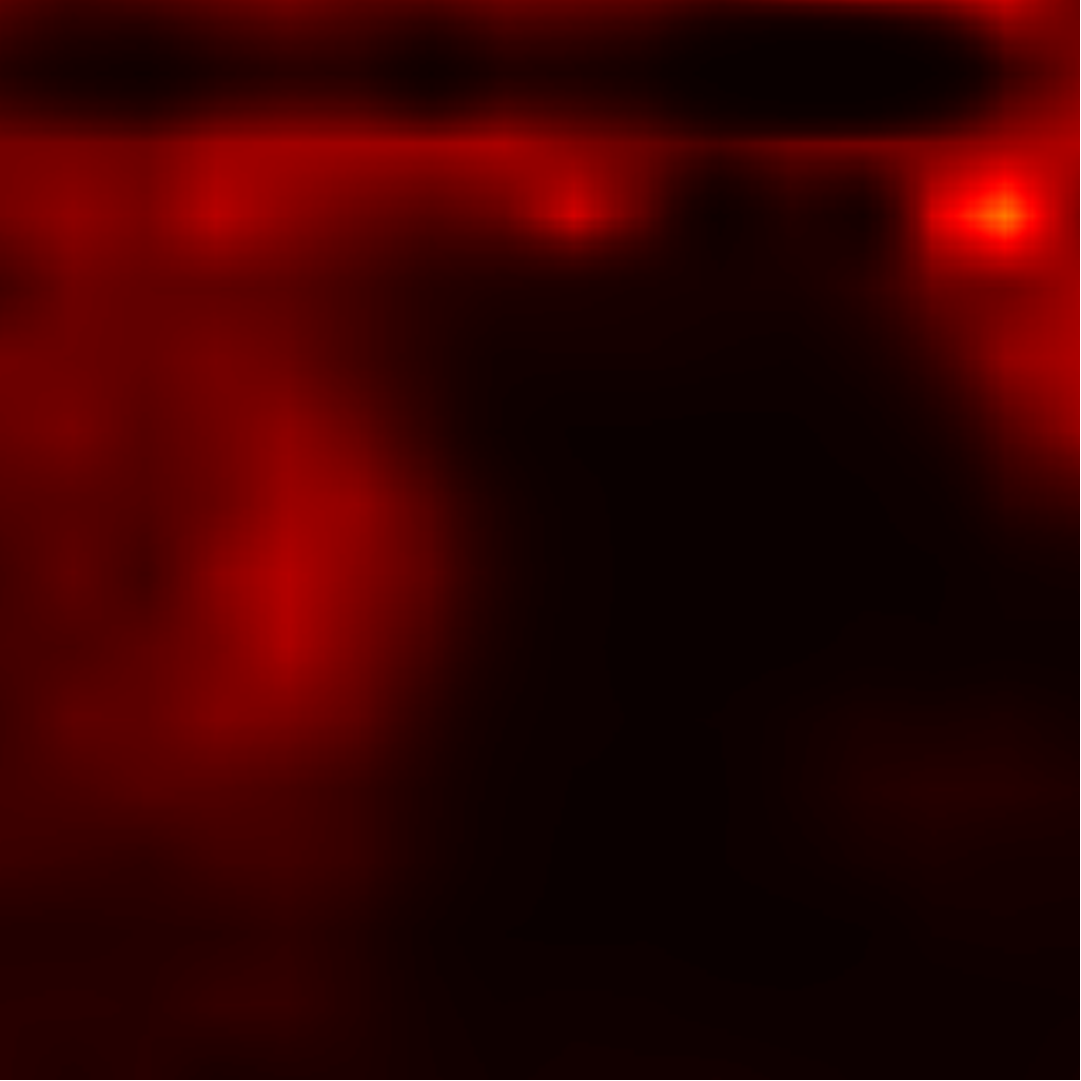}}%
\hfill
\subfloat[][RRDN (16x)\\ (\textbf{17.98} / \textbf{0.19} / \textbf{0.50} / 0.00)]{\includegraphics[width=0.24\linewidth]{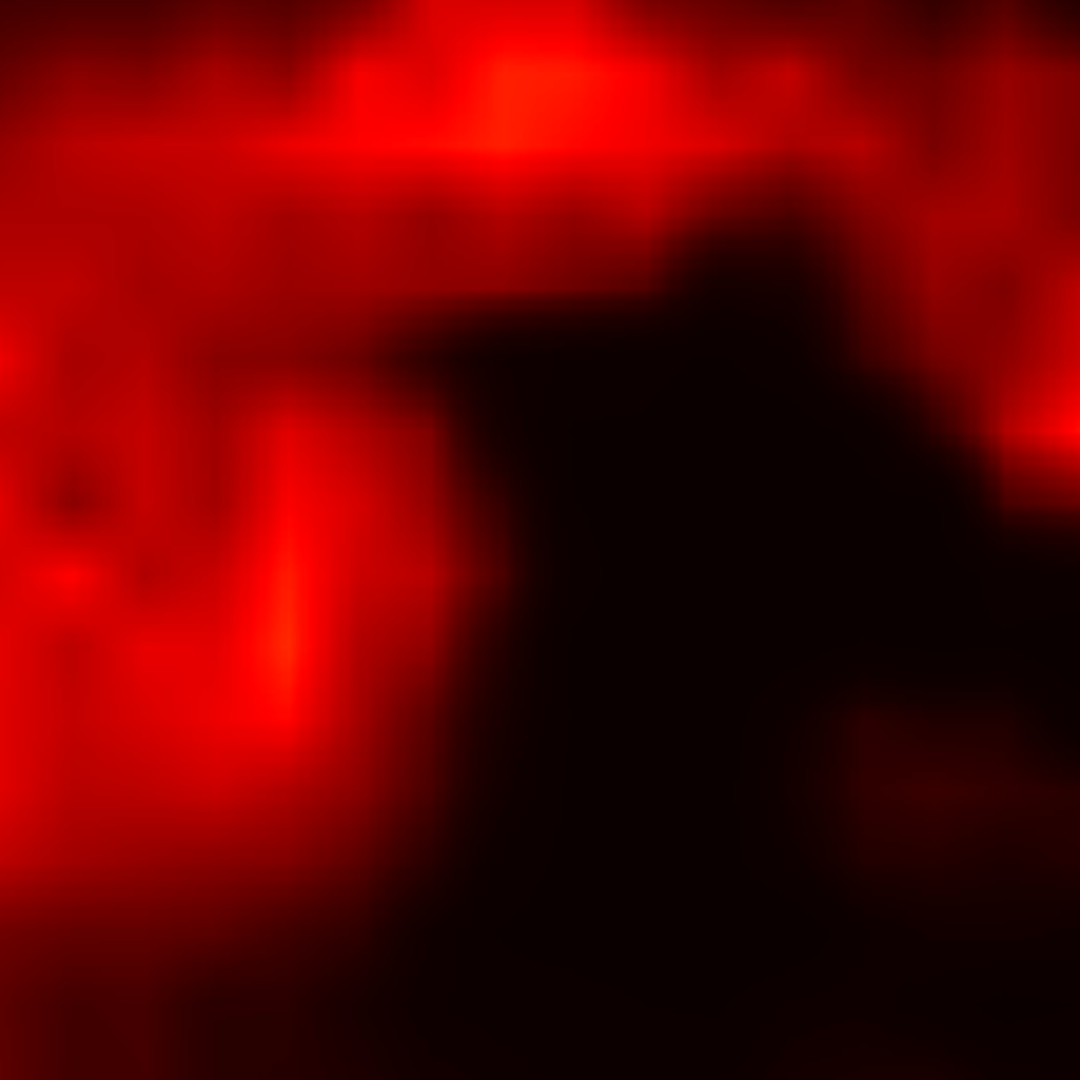}\label{fig:big_results_vegetation_RRDN}}%
\hfill
\subfloat[][Ours (16x)\\ (16.94 / 0.17 / 0.48 / \textbf{0.96})]{\includegraphics[width=0.24\linewidth]{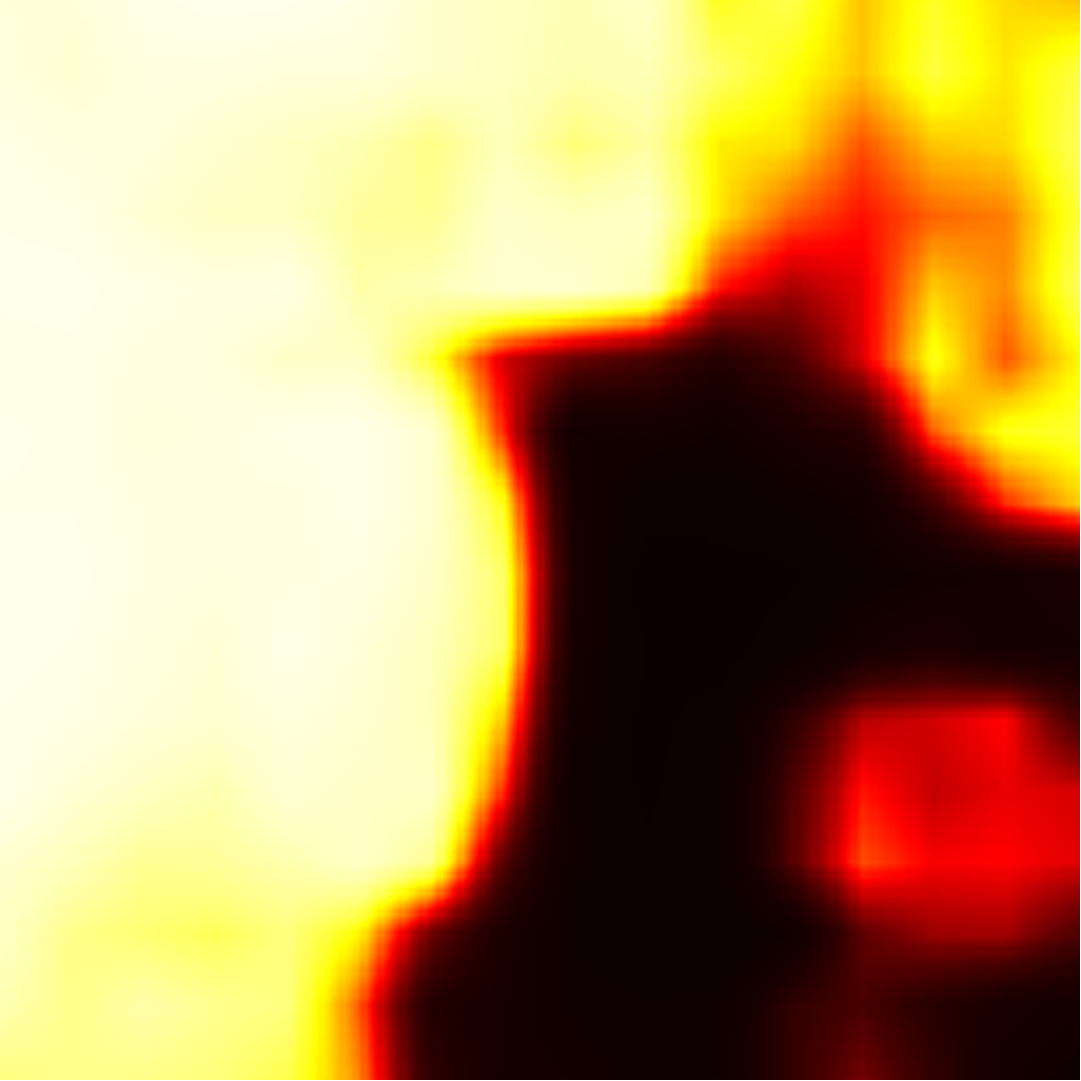}\label{fig:big_results_vegetation_GAN}}%
\hfill
\subfloat[][HR (16x)\\ ($\infty$ / 1.00 / 1.00 / 1.00)]{\includegraphics[width=0.24\linewidth]{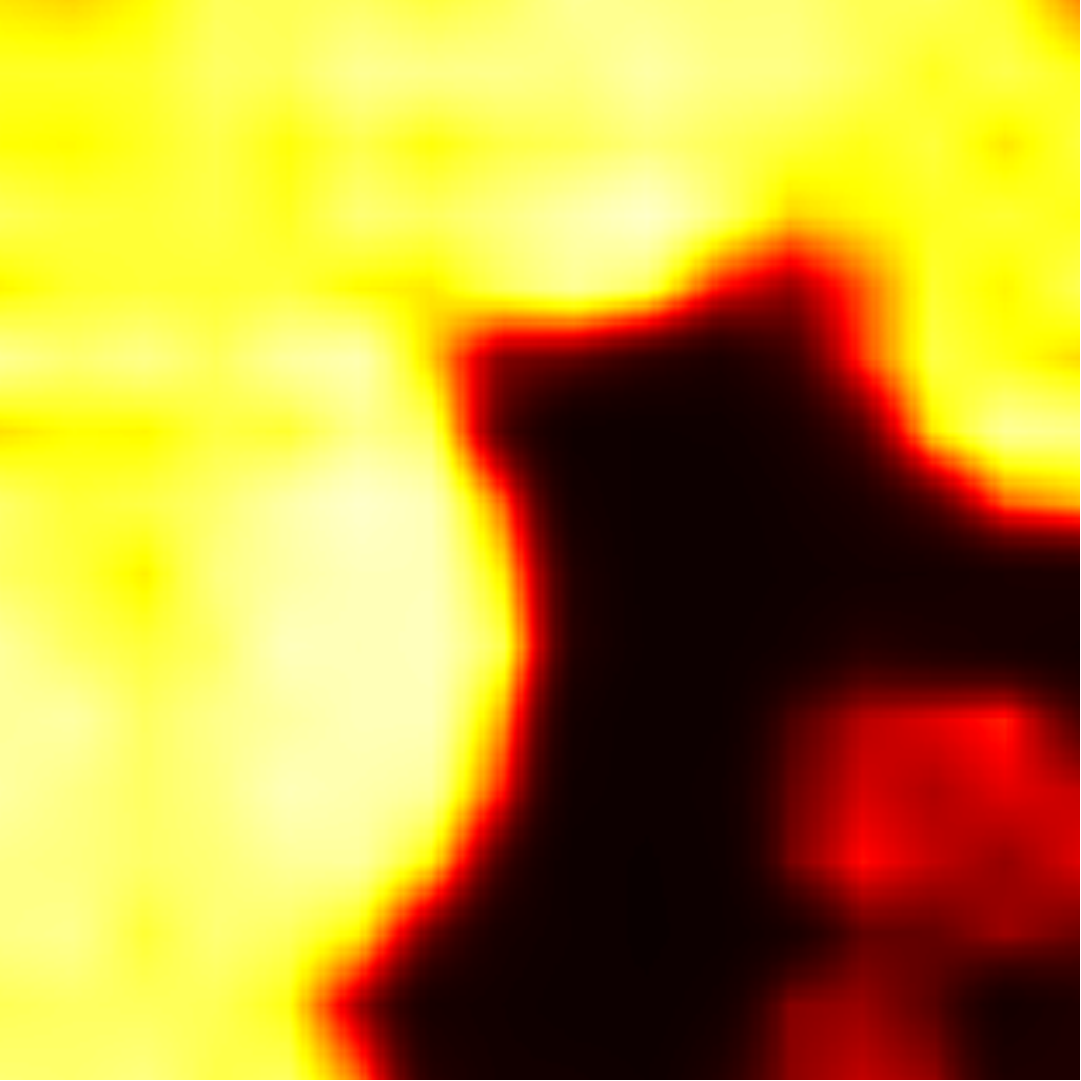}}%
\caption{Comparison of image super-resolution and semantic mask performance. We compare na\"ive bilinear upsampling, the state-of-the-art \gls{CNN}-based RRDN, the proposed \gls{GAN} approach, and the original high-resolution (HR) target image across four upsampling rates. For each example we report the PSNR, SSIM, MS-SSIM and IoU respectively. The proposed method shows the strongest perceptual fidelity and segmentation performance.}
\label{fig:big_results}
\end{figure}

\begin{figure}
  \centering
  \subfloat[]{\includegraphics[width=.33\linewidth]{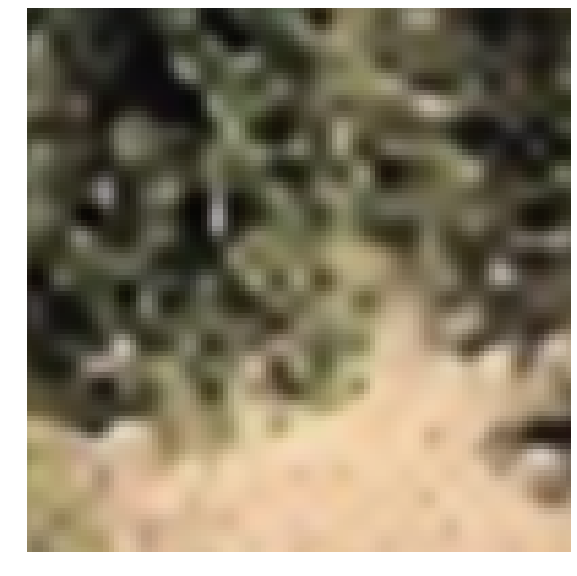}}
 \subfloat[]{\includegraphics[width=.33\linewidth]{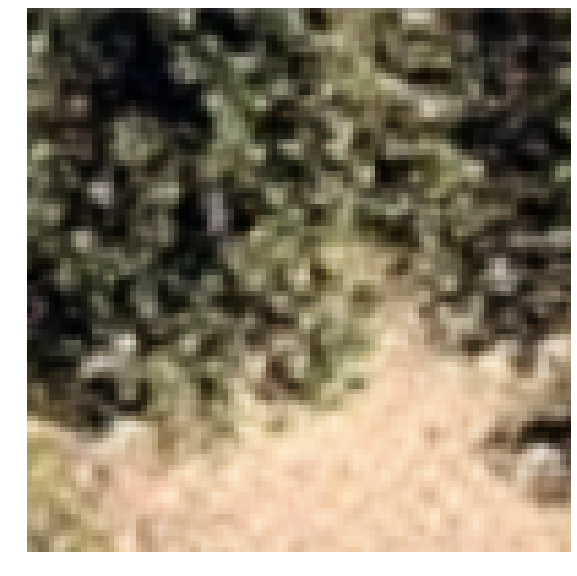}}
  \subfloat[]{\includegraphics[width=.33\linewidth]{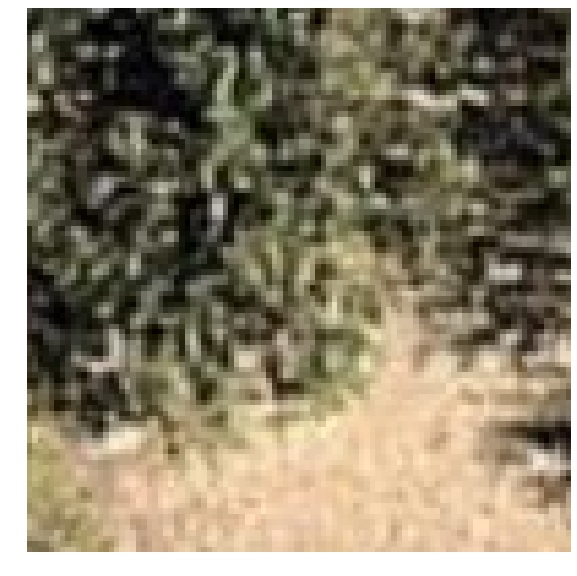}}
\caption{Detailed close-up of super-resolved imagery generated with (a) a \gls{CNN} trained with PSNR-based loss, (b) the proposed \gls{GAN}, and (c) the ground-truth original image. PSNR tries to maximise accuracy on average over pixels causing blurry reconstructions, whereas the \gls{GAN} is penalised for not recovering fine detail and hence learns to generate high-frequency textures.}
\label{fig:detail}
\end{figure}

We compared our approach to bilinear upsampling and RRDN, with typical results depicted in Fig.~\ref{fig:big_results}. 
Although not always quantitatively superior, images produced by the proposed \gls{GAN}-based approach are of superior perceptual quality, most notably in reconstruction of detail and texture. An additional detailed example is shown in Fig.~\ref{fig:detail}. It is understood that simple image metrics like PSNR and SSIM do not necessarily reflect qualitative fidelity \citep{Ledig2017Photo-realisticNetwork, Wang2019ESRGAN:Networks}, limiting their utility in assessing \gls{GAN} performance.

\subsection{Semantic Segmentation Performance}

We compared semantic segmentation masks derived from the low-resolution images, their high-resolution counterparts, and from upsampled images generated by the \gls{GAN}-based super-resolution models trained in the previous section. Qualitative and quantitative results are depicted in Fig.~\ref{fig:big_results}. Masks derived from the low-resolution imagery are either very poor or non-existent, while up-sampled imagery show significant improvement. The proposed \gls{GAN} provided the best performance, outperforming generator-only \gls{CNN} models (i.e.~RRDN) due to its ability to augment finer detail and textural patterns. This is clearest in Figs.~\ref{fig:big_results_vegetation_RRDN} and \ref{fig:big_results_vegetation_GAN}, with the latter showing dramatically improved \gls{IoU} compared with RRDN, increasing from 0 to 0.96 due to the generation of realistic tree foliage patterning which is clearly a crucial feature for the segmentation network. Note also that the semantic segmentation performance with the proposed method is very similar to performance using the ground-truth high-resolution images.

Table~\ref{tbl:miou} summarizes a comparison of semantic segmentation performance across the entire test set, for four upsampling factors. Each experiment starts with the same high-resolution images, then downsamples these by the specified factor. The 32$\times$ results are dealing with much less informative imagery than the the 4$\times$ imagery, explaining the lower absolute performance at higher upsampling rates.

The results show the proposed method significantly outperforming the baseline bilinear upsampling, and the CNN-based RRDN, with the performance margin widening for greater upsampling rates. At lower upsampling rates the proposed method yields an improvement of 11.8\% in \gls{mIoU} compared with the CNN-based method. At these scales the low-resolution images still contain sufficient detail that a CNN can essentially perform image sharpening to yield reasonable results. However, for higher upsampling rates image sharpening fails, and image segmentation performance for CNN-based methods is poor. The \gls{GAN}-based methodology reconstructs textural detail, yielding a 108\% improvement in \gls{IoU} over RRDN. 

\subsection{Class-Wise Analysis}

\begin{table}[b]
\caption{Comparison of \gls{mIoU} averaged across all classes. mIoU for the original high-resolution images is 0.543 and represents an upper bound on performance. Relative percentage improvement (Eq.~\ref{eq:perc_improvement}) shows substantial improvement in the proposed \gls{GAN} results compared with the \gls{CNN}-based approach.}\vspace{-1em}
\begin{center}
\begin{tabular}{cccc|c}
\hline
Scale             & Bilinear     & \gls{CNN}       & \gls{GAN} (ours)  & \% Improvement  \\ \hline \hline
4x                 & 0.389    &  0.468   & \textbf{0.523} & 11.8 \\ \hline
8x                 & 0.187    &  0.368   & \textbf{0.481} & 30.7 \\ \hline
16x               & 0.095   & 0.212    &  \textbf{0.388} & 83.0 \\ \hline
32x               & 0.068    &  0.120   & \textbf{0.250} & 108 \\ \hline
\end{tabular}
\end{center}
\label{tbl:miou}
\end{table}

\begin{figure}
\begin{center}
   \includegraphics[width=\linewidth,trim={0 0.5cm 0 1cm},clip]{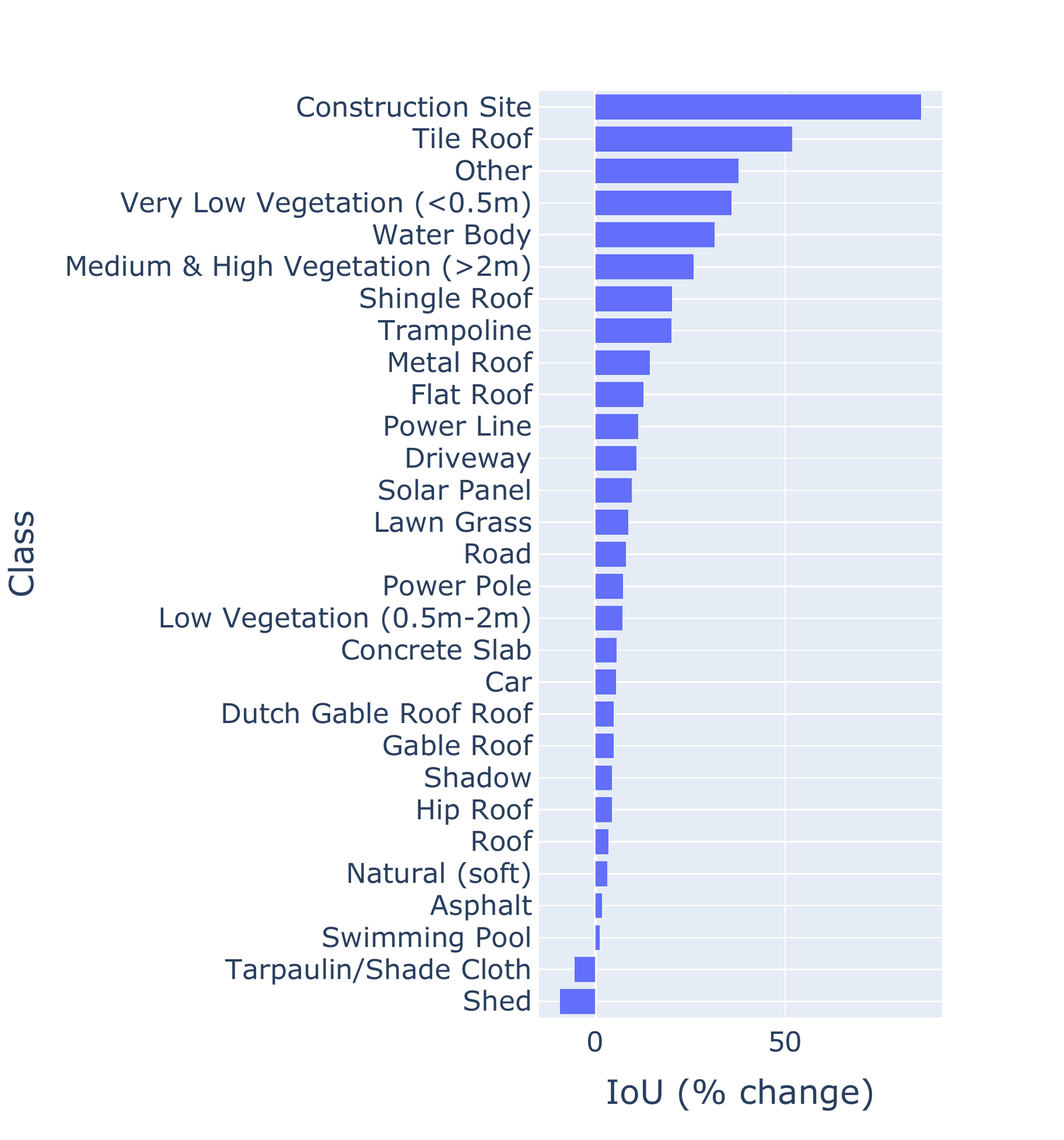}
\end{center}
   \caption{Percentage improvement (Eq.~\ref{eq:perc_improvement}) of class-wise mIoU between our \gls{GAN} approach and \gls{CNN} upsampling at 4x super-resolution. The proposed approach shows substantial performance gains of up to 86.0\%, showing the strongest improvement for classes with fine textural detail.}
\label{fig:4x_gan_cnn}
\end{figure}

To better understand the performance of each method, we compared the proposed approach with the CNN-based RRDN across individual semantic classes, for a conservative upsampling factor of 4$\times$. The results are depicted in Fig.~\ref{fig:4x_gan_cnn}, showing the proposed model outperforms RRDN in 27 out of 29 categories, with up to an 86.0\% improvement seen in the Construction Site class, and only small regressions in its weakest classes. %

The proposed approach provides the strongest advantage when resolving finely textured classes, e.g.~roofing (Tile Roof 52.0\%, Shingle Roof 20.3\% and Metal Roof 14.4\%) and vegetation (Very Low Vegetation 35.9\% and Medium-High Vegetation 26.0\%). The proposed method's use of both pixel-level and semantic fidelity allows it to infer and retain relevant textural detail. 

Similarly, the proposed method shows less of an advantage for categories with little textural detail, e.g. Asphalt and Shadow, and outline-oriented categories like roof shape (Dutch Gable, Gable, Hip Roofs). We expect these trends to be amplified for higher upsampling rates, following the results in Table~\ref{tbl:miou}.


\section{Conclusion and Future Work}
\label{sec_Conclusions}

We have demonstrated the benefits of super-resolution image enhancement for improving semantic segmentation performance in remote sensing applications. We proposed an end-to-end GAN-based framework that jointly optimised for image fidelity and semantic feature reconstruction, showing improved performance at both tasks. The proposed method significantly outperforms traditional and state-of-the-art CNN methods, both of which suffer from loss of texture and poor feature reconstruction. We achieved up to a 108\% improvement in IoU compared with a leading CNN-based method, pushing our achievable resolution factor as far as 32x whilst retaining meaningful results. 

This work advances deep learning-based image processing as a means of overcoming expensive and challenging hardware constraints in practical aerial imaging applications. As future work we envision jointly learning additional semantically meaningful tasks, and incorporating multiple views into semantically aware super-resolution.

\bibliographystyle{model2-names}

\end{document}